\PassOptionsToPackage{numbers,square}{natbib}
\documentclass[]{dreamvu_preamble/dreamvu}

\usepackage{lmodern}
\usepackage{amsmath,amsfonts,amssymb}
\usepackage{xspace}
\usepackage{url}
\usepackage{array}
\usepackage{tabularx}
\usepackage{adjustbox}
\usepackage{enumitem}
\usepackage{float}
\usepackage{nicefrac}
\usepackage{multirow}
\usepackage{booktabs}
\usepackage{xcolor}
\usepackage{tikz}
\usetikzlibrary{arrows.meta,positioning,calc}
\usepackage[capitalize]{cleveref}

\graphicspath{{figs/}}

\newcommand{\nbf}[1]{\noindent\textbf{#1.}~}
\newcommand{\groot}{GR00T N1.6\xspace}

\newcommand{\robobenchmart}{RoboBenchMart\xspace}

\title{SABER: A Scalable Action-Based Embodied Dataset for Real-World VLA Adaptation}

\author{DreamVu\protect\footnotemark[1]}

\abstract{
Robotic deployment in real-world environments depends on rich, domain-specific action data as much as on strong model architecture. General-purpose robot foundation models show modest performance in complex unseen tasks such as manipulation in a retail domain when applied out of the box. The root cause is a data gap: retail environments are structurally absent from general robot pretraining distributions, and the path to filling that gap through teleoperated in-store demonstrations is prohibitively expensive, logistically constrained, and difficult to scale.

We introduce SABER, a high-fidelity retail robotics action dataset built from over 100 hours of natural in-store capture across multiple real grocery environments. SABER is constructed from a dual-stream capture architecture designed to record human workers performing everyday retail tasks — stocking shelves, retrieving items, navigating aisles — in fully operational store conditions. Egocentric footage from head-mounted cameras worn by the human actors records fine-grained hand activity at the point of interaction, while exocentric 360° scene footage from DreamVu's ALIA camera [5] simultaneously observes all actors and activities across the entire space from a single fixed unit. This combination yields a uniquely complete picture of human retail behavior — dexterous hand activity, whole-body motion, spatial context, and scene dynamics — all captured without staging, scripting, or teleoperation overhead.

The resulting corpus contains 44.8K training samples organized across three complementary action representation streams. The first stream provides 25K latent action sequences derived from egocentric video via LAPA-style encoding, capturing the full distribution of manipulation behaviors without requiring proprioceptive annotations. The second stream provides 18.6K dexterous hand-pose trajectories extracted from egocentric video and retargeted to robot joint space, grounding fine manipulation skills in kinematically valid robot representations. The third stream provides 1.2K whole-body synchronized motion sequences derived from the ALIA exocentric view and retargeted to a humanoid embodiment, enabling locomotion and reach behaviors that span the full body. 

We apply SABER as a domain-specific post-training layer for modern Vision-Language-Action (VLA) systems. When applied to GR00T N1.6 via a shared-backbone multi-task post-training recipe trained jointly over all three supervision streams, SABER yields a mean success rate of \textbf{29.3\%} across ten retail manipulation tasks. This represents more than \textbf{2.19x improvement} over fine-tuning baselines (13.4\%), demonstrating that high-fidelity multimodal action supervision provides a critical path for domain-specific robot adaptation in real-world conditions. These results establish a clear finding: high-fidelity naturalistic human behavior — systematically captured, annotated, and retargeted — is a viable and scalable foundation for domain-specific robot adaptation. SABER demonstrates that the path to capable retail robots runs through better data, and that better data can be collected today, at scale, without a robot in the loop. The dataset and code are available at \url{https://dreamvu.ai/saber}.

\vspace{3mm}
}

\begin{document}
\maketitle

\footnotetext[1]{A detailed list of contributions and acknowledgments can be found in the Contributions section.}


\section{Introduction}
\label{sec:intro}

The deployment of autonomous robots in retail environments is one of the
most practically demanding problems in embodied AI. A grocery robot must
open refrigerators, retrieve products of varying shapes from crowded shelves,
pick items from the floor, load baskets, and do so amid visual clutter,
store-specific layouts, reflective packaging, occlusions, and continual
SKU variation. Unlike controlled laboratory benchmarks, retail deployment
depends on robust action priors under real operational variability.

\begin{figure}[t]
  \centering
  \includegraphics[width=1\textwidth]{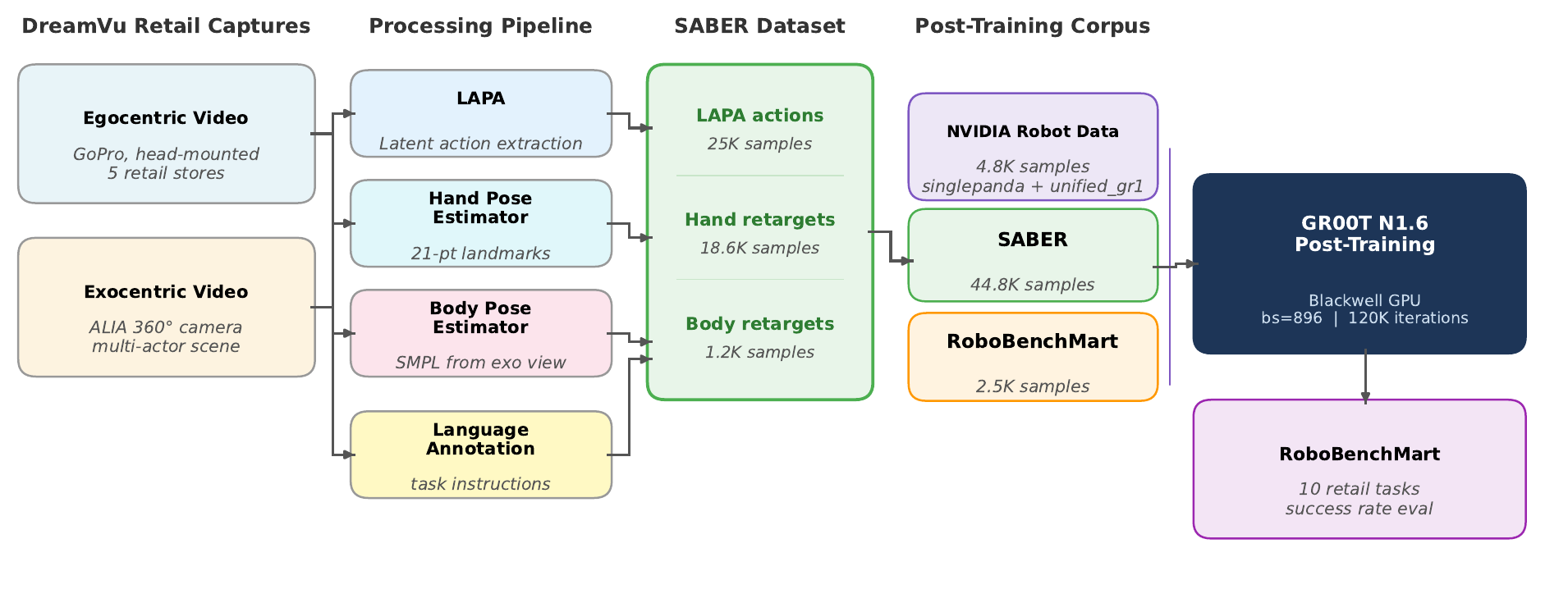}
  \caption{\small {\textbf{SABER: Three action representation streams.}
    From dual-camera in-store capture, three complementary supervision streams
    are derived: (1) LAPA latent action tokens (embodiment-agnostic motion);
    (2) dexterous hand-pose trajectories retargeted to robot joint space;
    (3) whole-body pose sequences retargeted to the Unitree G1 humanoid.
    All three streams share a VLM backbone and are trained with a joint
    multi-task objective. SABER-MM post-training achieves \textbf{29.3\%} mean success on ten \robobenchmart retail tasks.}}
  \label{fig:pipeline}
\end{figure}

A central bottleneck for achieving this is \emph{data}. Modern Vision-Language-Action (VLA) 
models have improved rapidly through large-scale pretraining on heterogeneous 
robot demonstrations~\cite{pi0,rt2,openvla,groot_n1}, but most available 
corpora remain limited in domain coverage and in the diversity of retail-relevant 
skills. This matters because store manipulation is not a narrow benchmark 
problem: it requires repeated exposure to articulated object interaction,
shelf reaches at multiple heights, grasping across packaging types, basket
loading, floor retrieval, and the embodied sequencing that connects these 
actions in realistic shopping workflows. When these skills are weakly 
represented in pretraining data, downstream performance collapses even 
for strong general models. NVIDIA's \groot~\cite{groot_n1} exemplifies 
this: a multi-embodiment humanoid model pretrained on large-scale 
teleoperated demonstrations across diverse robotic platforms.
Yet deployed directly on retail tasks from
\robobenchmart~\cite{robobenchmart}, its success rate is near zero.
This is not a model failure but a \emph{data failure}: the retail domain
is absent from general robot pretraining distributions, and acquiring
teleoperated retail demonstrations at meaningful scale is prohibitively
expensive.

We aim to demonstrate that domain-specific robot deployment needs a dedicated data layer.
In this paper, we introduce \textbf{SABER}, a high-fidelity retail robotics 
action dataset derived from DreamVu's in-store capture pipeline. SABER is built
from approximately 100 hours of in-store footage collected across multiple real
grocery stores---combining egocentric video from head-mounted GoPro cameras worn
by primary actors and exocentric 360$^\circ$ footage from DreamVu's ALIA
camera~\cite{DreamVuAlia2,DreamVuAlia}---and converted into $\sim$44.8K robot-training samples
spanning three complementary action representations: latent action tokens from
egocentric video, dexterous hand-pose trajectories retargeted to robot joint space
from egocentric video, and whole-body motion sequences retargeted to a humanoid
embodiment from the exocentric ALIA view. This construction is 
important for two reasons. First, it preserves the diversity and repetition of
real retail skills without requiring in-store robot teleoperation. Second, it 
supports multiple downstream use cases: domain adaptation for VLAs, action 
retargeting, dexterous manipulation learning, and humanoid whole-body control.

Within the broader foundation-model ecosystem, this paper highlights the role of
a domain-specific robotics action dataset for retail deployment, and demonstrates
that multimodal action supervision materially improves downstream policy learning.
To ground this claim empirically, we instantiate SABER as a post-training dataset
for NVIDIA's \groot~\cite{groot_n1} VLA. SABER-MM post-training uses a shared-backbone
multi-task objective over all three action representation streams (latent actions,
hand retargets, and body retargets), combined with a small amount of robot-native
anchor data and task-aligned seed data. This yields \textbf{29.3\%} mean success
on ten \robobenchmart~\cite{robobenchmart} tasks, compared with 13.4\% from baseline simulation-only fine-tuning—a \textbf{2.19x improvement} that demonstrates the effectiveness of multimodal action supervision for domain-specific robot adaptation.

Beyond immediate benchmark gains, datasets like SABER belong in the class of 
high-fidelity robotics action data for real-world, domain-specific deployment.
The underlying dataset is not limited to one embodiment or one training stack:
the same retail captures support latent action learning, hand retargeting, and
whole-body retargeting, making the corpus useful for a broader class of 
post-training and imitation pipelines.

\nbf{Contributions}
\begin{enumerate}[leftmargin=*,nosep]
  \item \textbf{A domain-specific retail robotics action dataset}: SABER provides
  $\sim$44.8K training samples derived from approximately 100 hours of real in-store
  human activity, with broad coverage of repeated retail skills such as shelf 
  interaction, articulated object manipulation, basket loading, and floor retrieval.
  \item \textbf{A multi-modal retargetable supervision pipeline}: from the same
  underlying captures, we derive latent actions, robot-space dexterous hand
  trajectories, and humanoid whole-body motion sequences, enabling both policy
  post-training and cross-embodiment retargeting.
  \item \textbf{A practical recipe for retail VLA adaptation}: we show how SABER 
  can be used as the dominant data source in a post-training mixture for a modern 
  VLA, achieving strong gains on retail manipulation tasks while preserving general
  manipulation priors.
  \item \textbf{A deployment-oriented framing}: we contextualize why real retail
  data is structurally different from generic robot data, and why scalable human
  capture can complement or outperform teleoperation-only collection for
  domain-specific robotic deployment.
\end{enumerate}

The dataset can be accessed via the following link \url{https://dreamvu.ai/saber}. A 10K-sample subset of SABER is released publicly under a CC BY-NC 4.0 license at \url{https://huggingface.co/datasets/DreamVu/SABER-10K}.

\section{Related Work}
\label{sec:related}

\begin{figure}[t]
  \centering
  \includegraphics[width=0.99\textwidth]{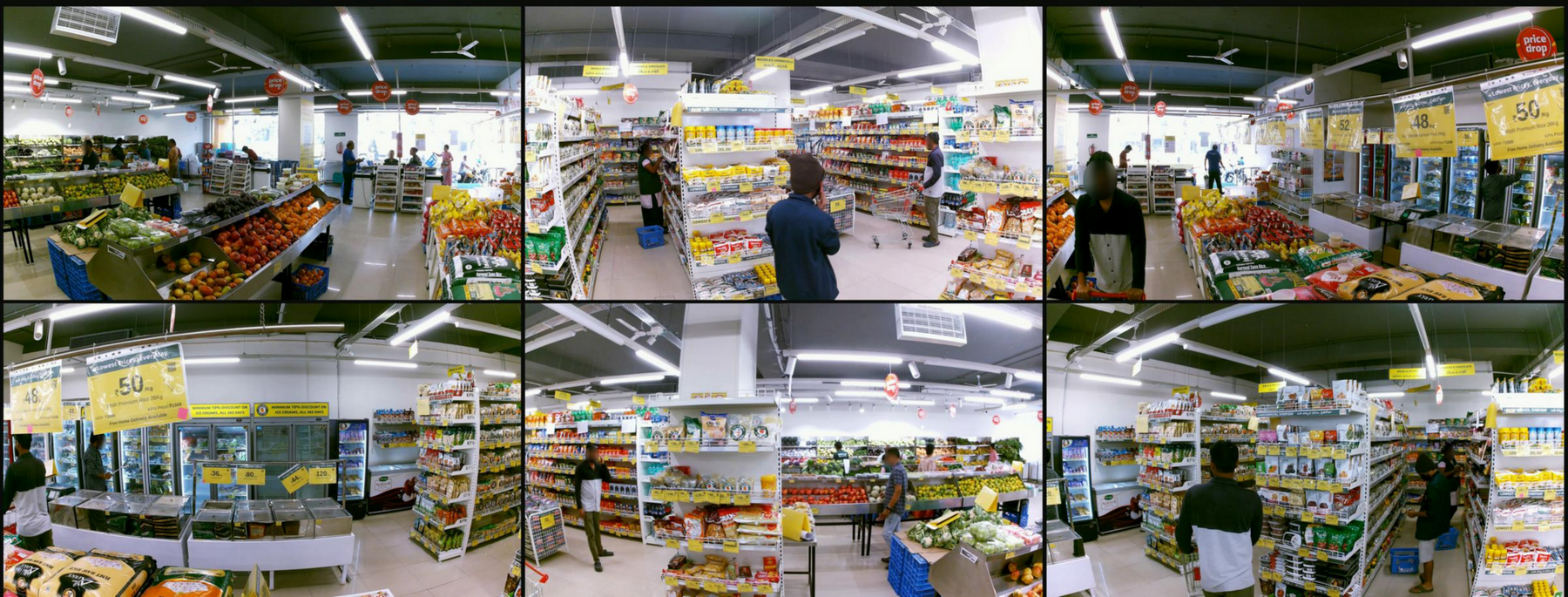}
  \caption{\textbf{ALIA 360$^\circ$ exocentric surround view.}
    Six calibrated wide-angle tiles from the DreamVu ALIA camera covering the
    full surround view of the store environment. The ALIA camera enables 
    simultaneous capture of all people in the
    scene from complementary angles, providing 360$^\circ$ visibility needed
    for robust SMPL pose estimation and Unitree G1 retargeting.}
  \label{fig:exo_surround}
\end{figure}

\nbf{Robot Foundation Models and VLAs}
Vision-Language-Action (VLA) models have rapidly advanced the state of robot learning by grounding manipulation policies in pretrained visual and linguistic representations. 
RT-2~\cite{rt2} demonstrated that VLM backbones can be fine-tuned to
produce robot actions, with language and vision grounding improving
generalization.
$\pi_0$~\cite{pi0} and OpenVLA~\cite{openvla} scaled this paradigm to
larger heterogeneous datasets, while \groot~\cite{groot_n1} and
Helix~\cite{helix} extended it to full humanoid embodiments using
dual-component architectures.
The field has converged on two dominant design patterns: single end-to-end
models (RT-2, OpenVLA, $\pi_0$) that process visual and language inputs in
a single forward pass, and dual-system architectures (GR00T N1, Helix) that
decouple a slower VLM reasoning component (System 2) from a faster
visuomotor flow-matching policy (System 1).
Open X-Embodiment~\cite{openx} aggregated 1M$+$ trajectories across 22
embodiments to study cross-embodiment generalization at scale.
LIBERO-PRO~\cite{libero_pro} quantified distribution-shift fragility:
models exceeding 90\% on standard LIBERO suites collapse to near zero
under modest perturbations to object positions or scene context---a result
that motivates domain-specific post-training rather than continued scaling
of general corpora.
DreamGen~\cite{dreamgen} and Cosmos-Reason1~\cite{cosmos_reason2} introduce
world-model-based data synthesis pipelines, generating plausible robot
demonstrations by rolling out learned visual dynamics---a complementary
direction to our human-video retargeting approach.

\definecolor{a0col}{RGB}{224, 92, 92}                                            
\definecolor{a1col}{RGB}{79,157,232}                                             
\definecolor{a2col}{RGB}{86,194,123}                                             
\definecolor{a3col}{RGB}{240,165,0}

\newcommand{\actionoverlay}[4]{%
    \begin{tikzpicture}[x=1pt,y=1pt,inner sep=1.5pt, font=\tiny\bfseries]          
      \node[fill=a0col!85, text=white, rounded corners=1pt] at ( 0,0) {$a_0$=#1};  
      \node[fill=a1col!85, text=white, rounded corners=1pt] at (24,0) {$a_1$=#2};  
      \node[fill=a2col!85, text=white, rounded corners=1pt] at (48,0) {$a_2$=#3};  
      \node[fill=a3col!85, text=white, rounded corners=1pt] at (72,0) {$a_3$=#4};  
    \end{tikzpicture}%
  }                                                                                
                  
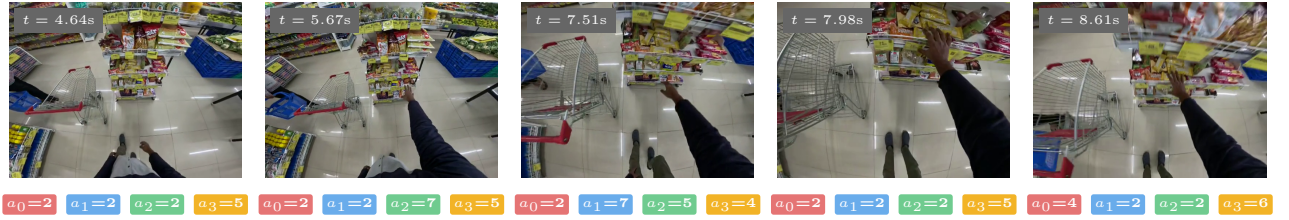
\begin{figure}[t]                                                                
    \centering    
    \setlength{\tabcolsep}{2pt}
    \begin{tabular}{ccccc}                                                         
      \begin{tikzpicture}                                                          
        \node[inner sep=0] (img)
  {\includegraphics[width=0.185\linewidth]{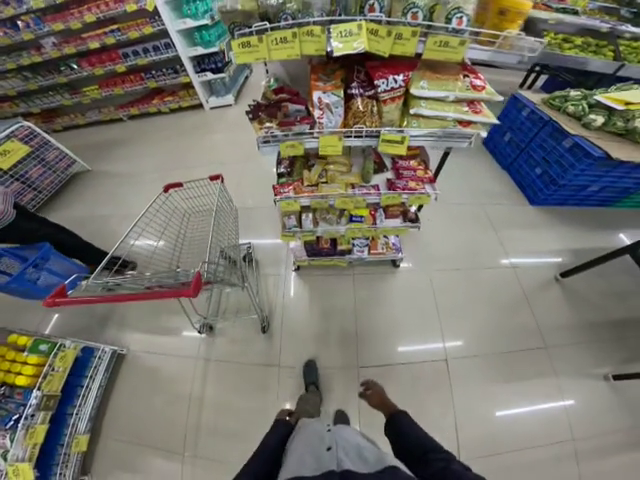}};                       
        \node[below right, inner sep=2pt] at (img.north west)
          {\colorbox{black!60}{\color{white}\tiny $t=4.64$s}};                     
      \end{tikzpicture} &                                                          
      \begin{tikzpicture}                                                          
        \node[inner sep=0] (img)                                                   
  {\includegraphics[width=0.185\linewidth]{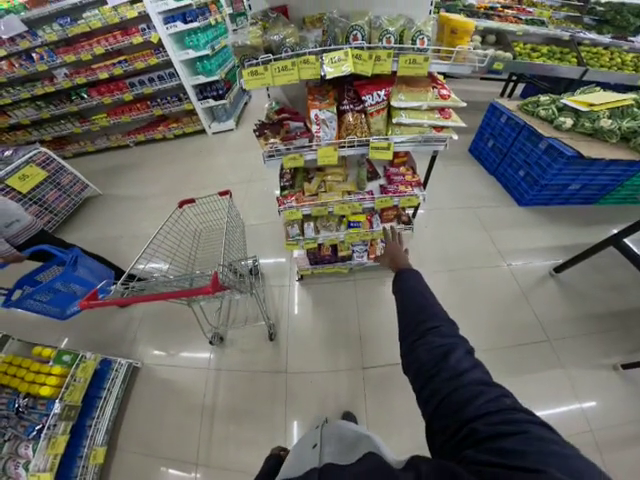}};                       
        \node[below right, inner sep=2pt] at (img.north west)
          {\colorbox{black!60}{\color{white}\tiny $t=5.67$s}};                     
      \end{tikzpicture} &                                                          
      \begin{tikzpicture}                                                          
        \node[inner sep=0] (img)
  {\includegraphics[width=0.185\linewidth]{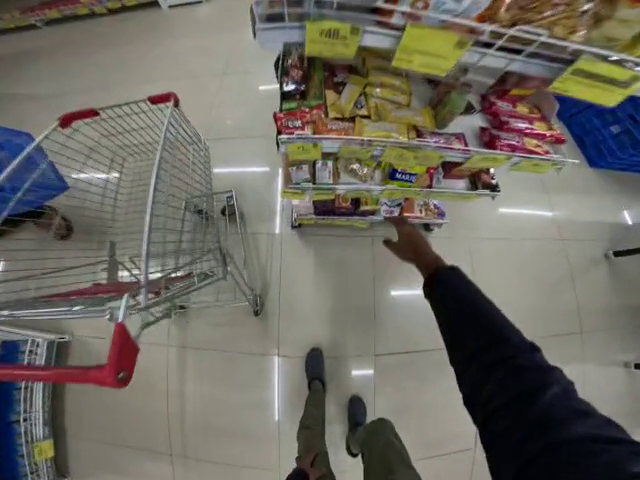}};                       
        \node[below right, inner sep=2pt] at (img.north west)
          {\colorbox{black!60}{\color{white}\tiny $t=7.51$s}};                     
      \end{tikzpicture} &
      \begin{tikzpicture}                                                          
        \node[inner sep=0] (img)
  {\includegraphics[width=0.185\linewidth]{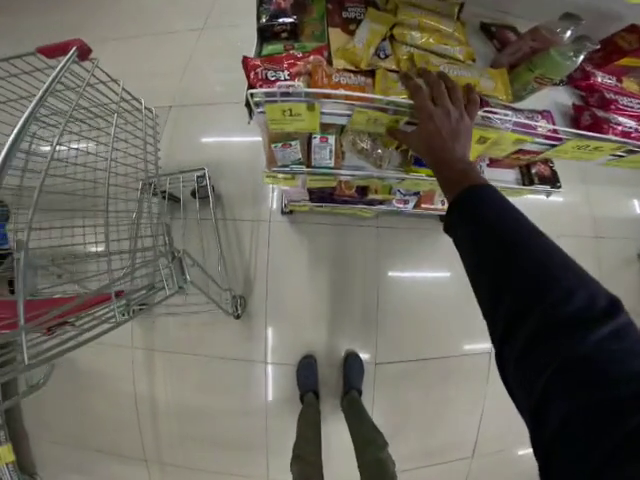}};                       
        \node[below right, inner sep=2pt] at (img.north west)
          {\colorbox{black!60}{\color{white}\tiny $t=7.98$s}};                     
      \end{tikzpicture} &                                                          
      \begin{tikzpicture}                                                          
        \node[inner sep=0] (img)                                                   
  {\includegraphics[width=0.185\linewidth]{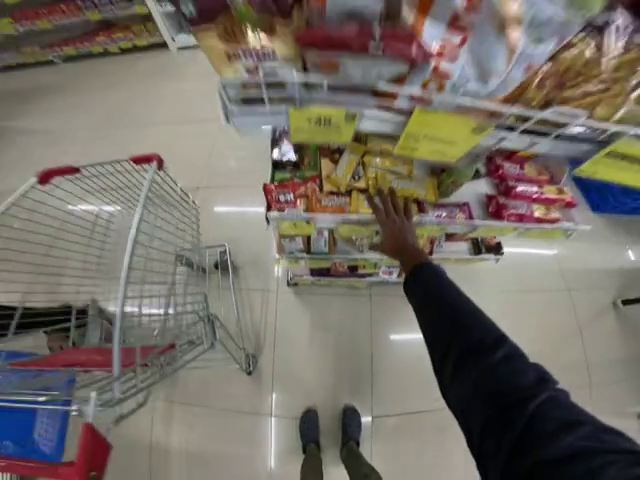}};
        \node[below right, inner sep=2pt] at (img.north west)                      
          {\colorbox{black!60}{\color{white}\tiny $t=8.61$s}};                     
      \end{tikzpicture} \\[2pt]
      \actionoverlay{2}{2}{2}{5} &
      \actionoverlay{2}{2}{7}{5} &                                                 
      \actionoverlay{2}{7}{5}{4} &
      \actionoverlay{2}{2}{2}{5} &                                                 
      \actionoverlay{4}{2}{2}{6} \\                                                
    \end{tabular}
    \caption{%
        The figure shows latent actions generated by LAPA-LAQ model \cite{lapa} at different frames of a video performing \emph{``Reach for snack packets on the lower shelves''}.  Each frame is annotated with its timestamp and the corresponding  discrete latent action tokens $a_0$--$a_3 \in \{0,\ldots,7\}$. These tokens capture discrete visual change between frames, irrespective of whether they were driven by hand motion or camera movement. Also, they are learned entirely without action labels, making them directly applicable to unlabeled human video.
    }
    \label{fig:ep041-frames}                                                       
  \end{figure}  

\begin{figure}[t]
  \centering
  \includegraphics[width=\textwidth]{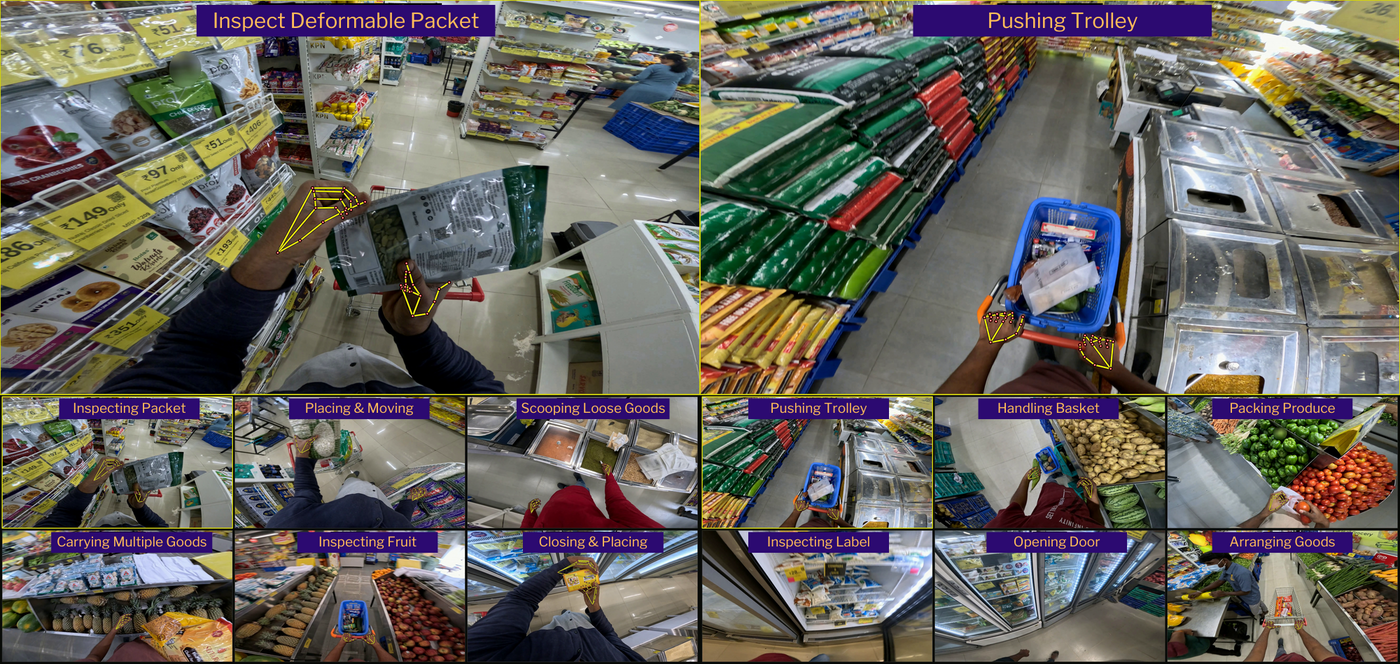}
  \caption{\textbf{Egocentric views across retail tasks.}
    Representative frames from the SABER egocentric capture corpus, spanning
    twelve core manipulation tasks: inspecting deformable packets, pushing trolleys,
    placing and moving items, carrying multiple items, inspecting fruits, scooping
    loose goods, opening and closing doors, inspecting labels, handling baskets,
    packing goods, and arranging goods. Each frame is captured from the head-mounted
    GoPro perspective of a primary actor during natural shopping activity across
    multiple store locations.}
  \label{fig:ego_collage}
\end{figure}

\nbf{Latent Action Learning from Video}
LAPA~\cite{lapa} learns a latent action space from unlabeled video via
an inverse dynamics objective over frame pairs $(I_t, I_{t+H})$,
producing pseudo-action tokens compatible with downstream policy training
without requiring robot joint labels. This is shown in Figure~\ref{fig:ep041-frames}.
JALA~\cite{jala} extends this line of work to joint-aligned latent
actions across embodiments.
SABER uses LAPA as one of three complementary supervision signals,
specializing it to egocentric retail video from head-mounted captures.
Unlike prior work that applies LAPA to generic web video, our captures
are domain-specific and paired with exocentric ALIA footage, enabling
richer per-episode context at annotation time.

\nbf{Human Pose Retargeting for Robot Control}
Transferring human kinematic structure to robot action spaces is an
active area of research.
Dex-Retargeting~\cite{qin2023anyteleop} propose AnyTeleop, an optimization-based retargeting framework that maps human hand keypoints extracted from egocentric RGB images to dexterous robot joint configurations without any learned, robot-specific models. Rather than training a dedicated retargeting network per embodiment, the method solves a per-frame optimization that preserves relative fingertip spatial relationships and hand-object interaction geometry while respecting joint limits and contact constraints. Given only a robot's URDF, the accompanying Dex-Retargeting plugin adapts to new arm-hand combinations without retraining, and a CUDA-accelerated, learning-free module handles collision avoidance. In contrast to prior systems such as DexPilot~\cite{handa2020dexpilot} and Telekinesis~\cite{sivakumar2022robotic}, which rely on hardware-specific training, this decoupling of retargeting from any particular robot or simulator enables generalization across diverse embodiments and environments within a single unified system.

SMPL-based~\cite{smpl} whole-body retargeting to humanoid platforms
maps 24-joint rotation vectors to robot-specific joint configurations,
enabling human motion capture to serve as robot demonstration data
without motion capture hardware.
SABER applies both pipelines to natural shopping video: Dex-Retargeting for
the 18.6K hand-pose stream extracted from egocentric footage, and
SMPL retargeting for the 1.2K whole-body stream extracted from
exocentric ALIA captures.

\nbf{Human Video and Cross-Embodiment Learning}
EgoMimic~\cite{egomimic} demonstrated that egocentric human video
improves robot policy generalization when combined with embodiment-aware
adaptation.
Ego4D~\cite{ego4d} and Ego-Exo4D~\cite{egoexo4d} established
egocentric video at scale as a resource for activity understanding from
both first- and third-person perspectives.
SABER extends this line of work by combining three representation
streams---latent actions, retargeted hand poses, and retargeted body
poses---trained jointly on a single shared backbone using a
stream-type conditioning embedding to route gradients to the
appropriate output distribution per stream.

\nbf{Retail and In-Store Robotics}
\robobenchmart~\cite{robobenchmart} is our primary evaluation benchmark,
comprising 10 manipulation tasks across fridge, board, floor, and
basket categories in a realistic retail environment.
SariBench~\cite{saribench} is a photorealistic 3D retail simulation
covering 250$+$ interactive grocery items;
its VLM-powered embodied agents achieve under 70\% success on easy tasks
versus near-perfect human performance---highlighting the gap between
general VLM capability and physical retail manipulation.
The AGIBOT World Challenge at ICRA 2026~\cite{agibot_world} offers a
\$530K competition spanning shelf stocking, barcode scanning, and
logistics tasks, further confirming industry interest in retail-specific
embodied benchmarks.
RoboMIND~\cite{robomind} provides 107K manipulation trajectories across
multiple embodiments with an explicit retail category, currently the
largest public manipulation dataset with in-store coverage.
On the industrial deployment side, Sereact Cortex~\cite{sereact_cortex}
achieves $\approx$98\% success on warehouse pick-and-place at Rohlik
Group's e-grocery fulfillment centers, using a Mixture-of-Experts VLA
trained on proprietary deployment data from 100$+$ industrial sites.
Cortex 2.0 integrates a latent world model for long-horizon failure
recovery, foreshadowing the direction that production retail VLA systems
are taking.
Despite this progress, no existing work provides a large-scale dataset
of \emph{in-store} human manipulation video with multi-modal kinematic
annotation targeting humanoid embodiments.
SABER addresses this gap directly: the dataset is built from unpaired
human shopping video captured with consumer hardware (GoPro head-mount
and DreamVu ALIA 360° camera), enriched with three complementary
kinematic supervision signals, and evaluated on a realistic retail
manipulation benchmark designed for real-world VLA deployment.

\section{Why Retail Demands Domain-Specific Data}
\label{sec:gap}

\begin{figure}[t]
  \centering
  \includegraphics[width=0.97\textwidth]{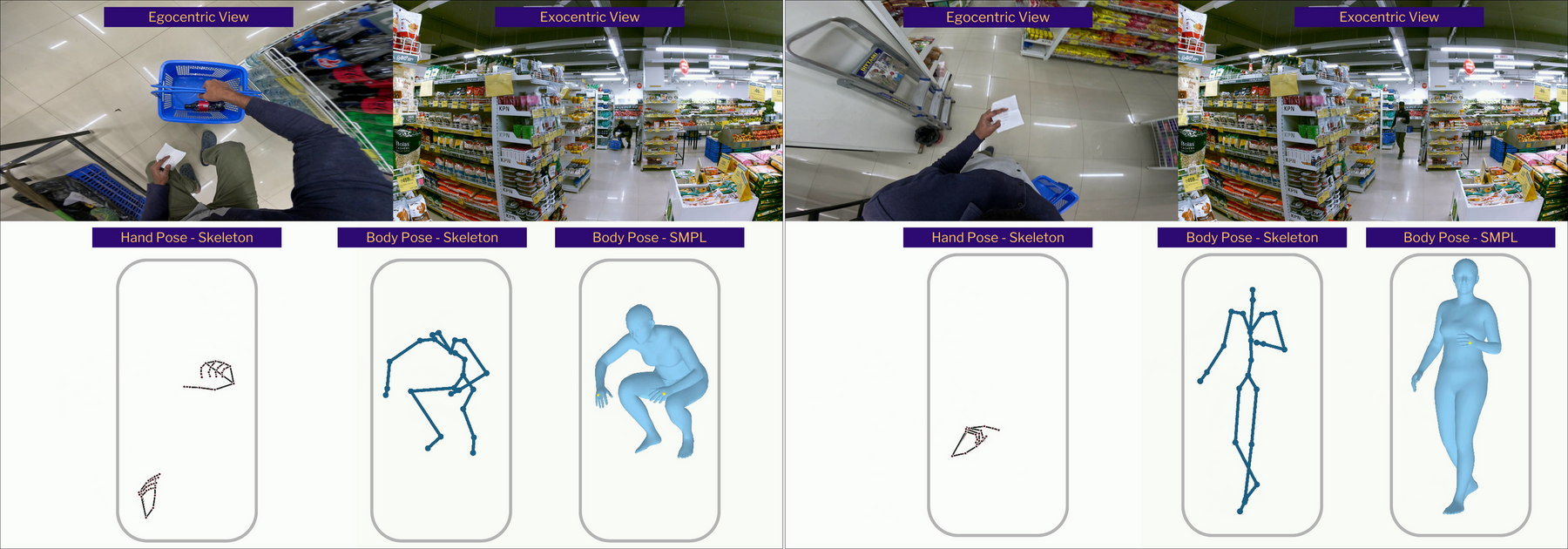}
  \caption{\textbf{SABER data collection and modalities.}
    Each capture session uses two simultaneous camera systems: a head-mounted
    GoPro (egocentric, primary actor view) and a DreamVu ALIA 360$^\circ$
    camera (exocentric, full-scene view of all people).
    Streams 1 and 2 (LAPA latent tokens and Dex-Retargeting hand retargets) are derived
    from the egocentric footage; Stream 3 (SMPL whole-body poses retargeted to
    Unitree G1) is derived from the exocentric ALIA footage.
    No robot hardware is present in any capture.}
  \label{fig:modalities}
\end{figure}

Retail robotics presents three structural challenges absent from standard manipulation benchmarks.
First, the \emph{skill distribution} is distinct: articulated object interaction, multi-height
shelf reaching, basket loading, floor retrieval, and context-dependent placement, all executed
repeatedly across hundreds of SKUs. Second, \emph{long-tail scene variation}: dense shelves,
restocking, occlusions, varied lighting, and product deformability create real-world
complexity that generic datasets cannot approximate. Third, \emph{repetition matters}: a
model must see skill families repeatedly across contexts (grasping bottles from different
shelf heights, opening fridges from varied approach angles) to achieve reliable deployment.

Three primary approaches exist for collecting domain-specific robot data, each with
distinct tradeoffs:
\begin{itemize}[leftmargin=*,nosep]
  \item \textbf{Teleoperation}: High fidelity (real dynamics, true embodiment) but slow
    acquisition; in-store operation is operationally disruptive and costly.
  \item \textbf{Simulation}: Fast and cheap but suffers from sim-to-real gaps in contact-rich
    tasks; data is clean and unrealistic. Embodiment-locked unless retargeted.
  \item \textbf{Human video retargeting}: Real-world scale, embodiment-agnostic, naturally
    diverse—but data quality depends entirely on pose annotation accuracy.
\end{itemize}

Under ideal conditions (perfect engines, unlimited time, embodiment match), the efficiency
ranking would be dominated by simulators because they can scale much better than human videos
or teleoperation. As of today, however, the fidelity ranking is inverted - simulation suffers
from a significant domain gap with real deployment, and teleoperation is prohibitively expensive. Human video retargeting
offers a practical middle path: real data at scale with flexibility, provided that the
quality of action estimation and annotation is maintained. 
SABER operationalizes this approach by deriving ground-truth
annotated action representations from \emph{$\sim$100 hours of natural in-store human
capture}, enabling high-fidelity domain-specific training.

\section{The SABER Dataset}
\label{sec:data}

\subsection{Data Collection}

\begin{figure}[t]
  \centering
  \includegraphics[width=0.95\textwidth]{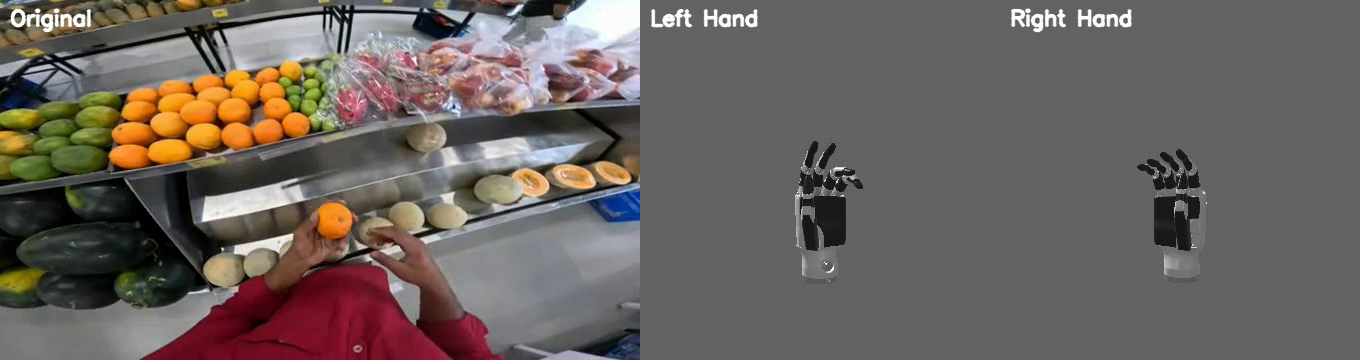}
  \caption{\textbf{Hand pose retargeting.}
    \textit{Left}: egocentric source frame showing a grasp-and-inspect
    action during natural shopping.
    \textit{Right}: corresponding left and right hand poses retargeted
    to robot dexterous hand joint space via Dex-Retargeting~\cite{qin2023anyteleop},
    preserving contact constraints and grasp geometry.
    These retargeted trajectories form Stream 2 of the SABER dataset.}
  \label{fig:hand_retargeting}
\end{figure}

DreamVu's retail captures span multiple real grocery store locations across diverse 
layouts, lighting conditions, and product assortments. The collection strategy is 
driven by skill coverage rather than passive recording alone: actors perform the full 
shopping workflow, including aisle navigation, browsing, repeated shelf interaction, 
picking and placing products, handling baskets and carts, opening and closing refrigerators,
and retrieving items from the floor. This emphasis on both \emph{variety} and 
\emph{repetition} is central to SABER's value as a robotics dataset.

SABER captures are collected using two complementary camera systems deployed
simultaneously in each session, illustrated in \cref{fig:modalities}.

\nbf{Egocentric view}
Primary actors wear a head-mounted GoPro camera recording at 480p from the
first-person perspective.
This view keeps hand-object interactions prominently in frame, making it
well-suited for extracting hand motion and grasp dynamics via LAPA and Dex-Retargeting.
Streams 1 (LAPA latent actions) and 2 (hand retargeting using Dex-Retargeting) are derived
from this egocentric footage.

\nbf{Exocentric view}
A DreamVu ALIA omnidirectional camera~\cite{DreamVuAlia2,DreamVuAlia} provides a fixed
360$^\circ$ view of the scene that simultaneously captures all people present,
including the primary actor(s).
The camera produces six calibrated and synchronized wide-angle views that span the full
surround environment of the store environment, as shown in \cref{fig:exo_surround}.
The full-scene perspective enables robust whole-body pose estimation across the
entire skeleton---including legs, torso, and head---which is not reliably visible
from the egocentric view alone.
Stream 3 (body pose retargets) is derived exclusively from this exocentric footage.

The egocentric and exocentric views shown in \cref{fig:exo_surround} and \cref{fig:ego_collage}
represent the complementary perspectives captured during SABER collection.
From approximately 100 hours of synchronized in-store activity across both cameras,
we construct a post-training corpus that preserves real deployment context while
remaining compatible with multiple downstream action representations.

\subsection{Three Action Representation Streams}
\label{sec:streams}

SABER derives three complementary representations from its dual-camera
corpus---egocentric for Streams 1 and 2, exocentric for Stream 3---each
capturing a different level of kinematic abstraction (\cref{fig:modalities}).

\nbf{Stream 1 --- LAPA Latent Actions (25K episodes)}
A total of 25K egocentric episodes are processed through the LAPA~\cite{lapa}
pipeline. Each episodes spans an approximate duration of ten seconds. 
For each frame pair at a fixed window size H $(I_t, I_{t+H})$ in the episode, an inverse dynamics
encoder maps the observed visual transition to a compact latent action
token $z_t$.
No robot joint labels are required: the token is self-supervised to
encode precisely what caused the visual change between frames, capturing
whole-arm motion, reach trajectories, and grasping dynamics at a
semantic level.
LAPA tokens are directly usable by downstream latent-action policies, and in our
experiments they align naturally with the target VLA action interface.

\nbf{Stream 2 --- Dex-Retargeting for Hand Retargeting (18.6K episodes)}
For all 18.6K egocentric episodes, a 21-point hand landmark skeleton is first
estimated for each frame using a hand pose estimator. A team of trained annotators then reviews and corrects
these automated estimates frame-by-frame, ensuring anatomically accurate and
contact-consistent hand configurations.

To validate annotation quality, we employ a rigorous multi-stage QC process:
(1) annotators receive standardized training on anatomical hand structure and
contact identification; (2) all 18.6K episodes undergo an annotate$\to$QC$\to$re-annotate
cycle in which a supervising annotator flags anatomically invalid frames and
inconsistent contact transitions; (3) flagged frames are re-corrected by the
original annotator with supervisor feedback; (4) inter-annotator agreement
is measured by having independent annotators re-review 5\% of episodes.
This hybrid approach---combining fast automated initialization with iteratively-refined
human ground-truth---ensures that hand joint annotations are genuinely ground truth
rather than noisy automated estimates.

\begin{figure}[H]
  \centering
  \includegraphics[width=0.92\textwidth]{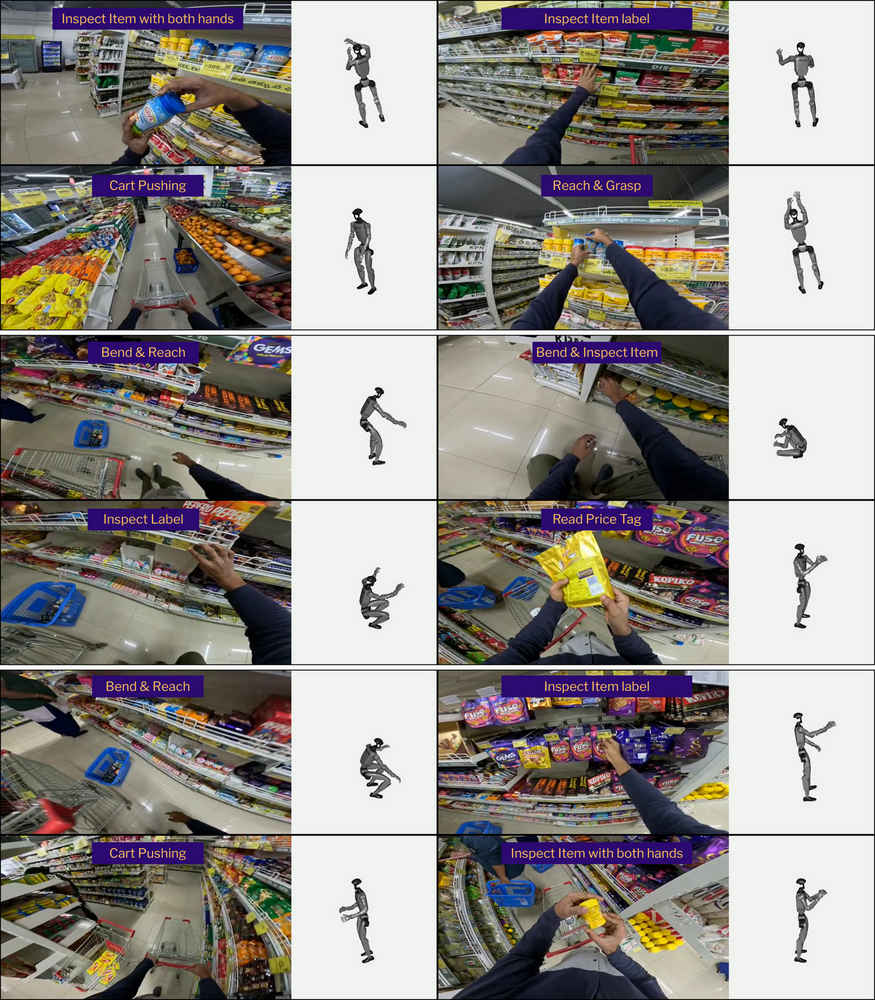}
  \caption{\textbf{Whole-body pose retargeting to Unitree G1.}
    Multiple instances of egocentric to whole-body retargeting. Each row pair shows:
    \textit{Left}: egocentric source view from the store capture session.
    \textit{Right}: corresponding whole-body pose retargeted to the Unitree
    G1 humanoid joint configuration, derived from SMPL parameters estimated
    from the ALIA exocentric view of the same session.
    These retargeted motion sequences form Stream 3 of the SABER dataset.}
  \label{fig:body_retargeting}
\end{figure}

Dex-Retargeting~\cite{qin2023anyteleop} then solves a constrained optimization to retarget
these corrected human hand configurations to the target robot's dexterous
hand joint angles, preserving contact constraints and hand-object interaction
geometry.
The result is a per-episode trajectory of robot-space hand joint angles
derived from high-fidelity, iteratively-validated hand pose data.
This representation provides explicit, geometrically precise supervision
for fine-grained finger control---the level of detail that LAPA latent
tokens encode implicitly and weakly.
\cref{fig:hand_retargeting} shows a representative frame from the
retargeting pipeline alongside the corresponding robot hand poses.

\nbf{Stream 3 --- Body Pose Retargets to a Humanoid (1.2K episodes)}
A subset of 1.2K episodes with synchronized ALIA exocentric footage are
processed through a whole-body pose annotation pipeline.
The DreamVu ALIA 360$^\circ$ camera~\cite{DreamVuAlia2,DreamVuAlia} provides a full-scene
view that captures the body-pose of all the people in the scene -including legs and
torso that may be otherwise occluded or out-of-frame in the egocentric view - making it
the appropriate source for full-body SMPL estimation.
SMPL~\cite{smpl} parameters (24-joint rotation vectors and 10-dim shape
coefficients) are estimated per frame using a specialized human pose
estimation framework. Trained annotators then thoroughly review and correct these
estimates frame-by-frame to ensure anatomical validity and temporal consistency.

Annotation quality for body pose is particularly critical because retargeting fidelity
depends on precise shoulder, hip, and knee angles. We enforce quality through a
structured validation pipeline: (1) annotators complete a training module
covering anatomical constraints (2) all 1.2K episodes
undergo iterative annotate$\to$supervise$\to$re-annotate cycles in which ambiguous frames
(occlusions, rapid motion) are flagged and re-examined; (3) inter-annotator agreement is
measured on a held-out 10\% of episodes by having an independent annotator re-mark poses.

The human-corrected SMPL parameters are then retargeted to the Unitree G1 humanoid~\cite{unitreeg1}
joint configuration using an SMPL-to-robot retargeting procedure that maps
human joint rotations to G1-compatible joint angles while respecting kinematic
limits.
The retargeted trajectories provide whole-body coordination supervision:
weight shifts, reach-and-step sequences, and the full torso-arm coordination
required for floor retrieval and extended-shelf tasks.
The smaller count (1.2K vs.\ 18.6K) reflects the annotation cost of frame-by-frame
human review for full-body pose data---a non-trivial investment that ensures
these trajectories represent genuine ground truth validated through multi-stage QC.
Despite this smaller scale, body pose retargets provide a qualitatively distinct
supervision signal absent from both LAPA tokens and Dex-Retargeting hand retargets:
torso-arm-leg coordination at the specific joint resolution of the target
humanoid platform.
\cref{fig:body_retargeting} illustrates the retargeting pipeline from
exocentric source footage to Unitree G1 joint configuration.

\subsection{Dataset Statistics}

\begin{figure}[t]
  \centering
  \includegraphics[width=\textwidth]{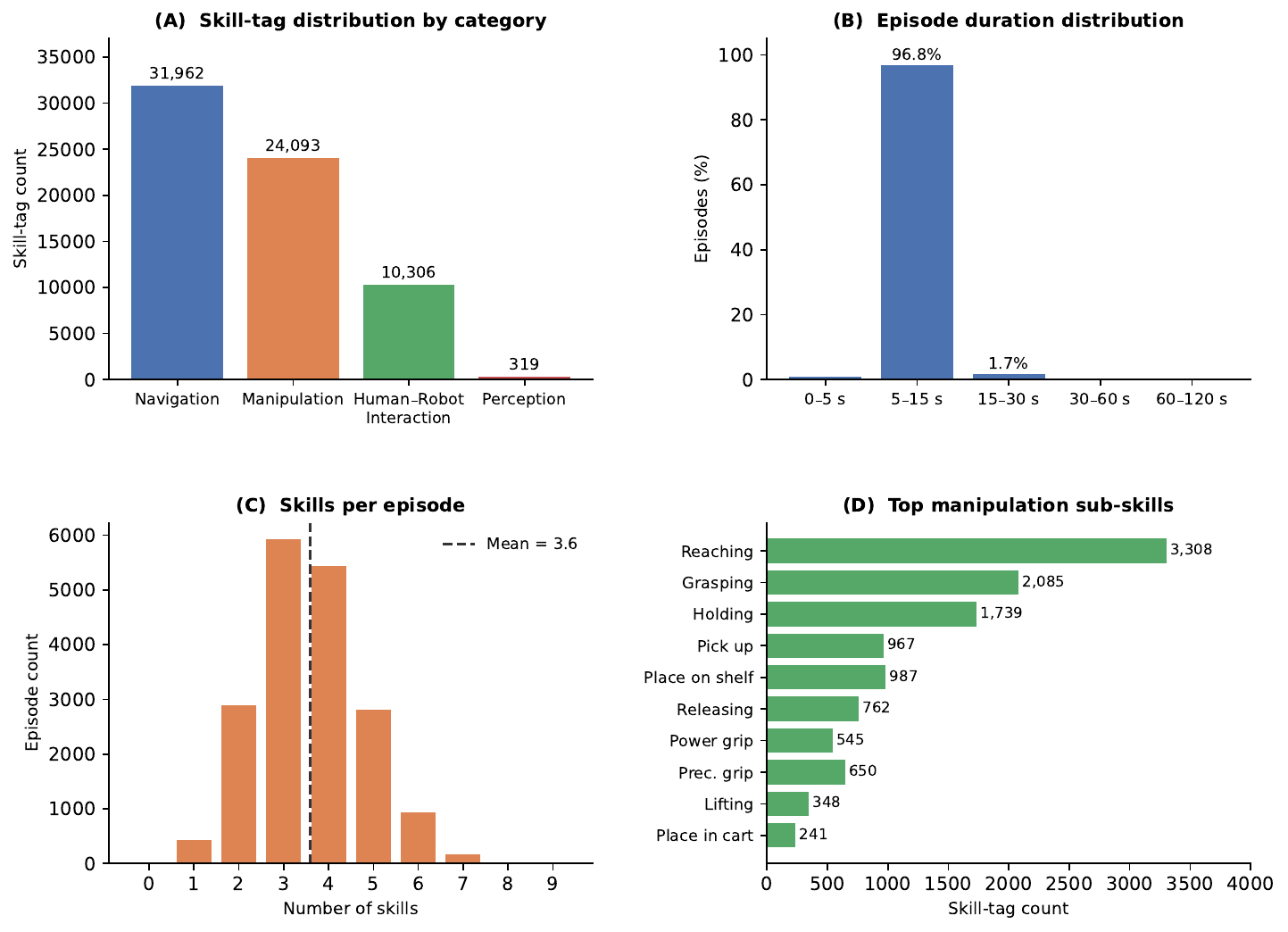}
  \caption{\textbf{DreamVu retail capture database statistics.}
    (A)~Skill-tag counts across four main categories; navigation and
    manipulation together account for 84\% of annotations.
    (B)~Episode duration distribution: 94\% of episodes fall in the
    5--15\,s window.
    (C)~Skills per episode, with mean 3.6 (dashed line)---most episodes
    encode 3--4 natural action transitions.
    (D)~Top manipulation sub-skills by count; multiple grasp types
    (power, precision, pinch) are all well-represented.}
  \label{fig:dataset_stats}
\end{figure}

\cref{fig:dataset_stats} summarises the DreamVu retail capture database
that underlies SABER.
Statistics are derived from the raw captures (egocentric and exocentric)
before action extraction and retargeting, thereby characterising the behavioural richness of the source material.

\nbf{Skill-tag distribution (Panel A)}
Across the corpus, 66,680 skill-tag annotations span four main
categories: \emph{navigation} (31,962; 47.9\%), \emph{manipulation}
(24,093; 36.1\%), \emph{human--robot interaction} (10,306; 15.5\%), and
\emph{perception} (319; 0.5\%).
The near-equal split between navigation and manipulation reflects the
natural shopping workflow: actors travel between shelf zones
(\emph{navigation}) and then interact with products
(\emph{manipulation}).
This balance means SABER encodes not only isolated grasp events but also
the approach trajectories and context that precede them---a richer
signal than laboratory pick-and-place datasets that begin from a
pre-posed arm state.

\nbf{Episode duration (Panel B)}
94\% of episodes fall in the 5--15\,s window, with the distribution sharply 
concentrated around this range. These clips are long enough to capture complete 
manipulation primitives (approach, contact, release) but short enough to avoid 
the noise introduced by unrelated multi-task transitions within a single clip, 
making them well-suited for action-model post-training. Considering the consistency of the SABER dataset it can also be used for pre-training stage.

\nbf{Skills per episode (Panel C)}
The distribution of skill-tags per episode peaks at 3--4 skills
(mean 3.6), indicating that most captures encode a small, coherent
sequence of actions rather than a single isolated primitive.
This multi-skill structure is valuable for training: the model observes
natural action chaining (e.g., reaching $\to$ grasping $\to$ placing)
rather than only isolated motion segments.

\nbf{Manipulation sub-skill coverage (Panel D)}
Within the manipulation category, the top sub-skills by count are
reaching (3,308), grasping (2,085), holding (1,739), picking up items
(967), placing on shelf (987), releasing (762), power grip (545),
precision grip (650), lifting (348), and placing in cart (241).
The diversity of grasp types---power, precision, and pinch grips all
present at scale---is directly relevant to SABER's Dex-Retargeting stream: the
retargeted hand-pose trajectories cover a realistic distribution of
real-world grasp configurations rather than a narrow laboratory
repertoire.
SABER comprises 44.8K total samples drawn from 25K LAPA latent actions, 18.6K Dex-Retargeting hand-pose retargets, and 1.2K body pose retargets, all processed from the underlying capture database.

As described earlier, the dataset spans diverse manipulation tasks and . is collected across multiple stores under diverse lighting conditions and varying shelf layouts with different product assortments. Complementing visual observations, the dataset includes language instructions for each episode covering product names, shelf positions, and action types. All data are formatted as LeRobot-compatible triples consisting of observation frames, action representations, and language instructions, with stream type encoded as additional metadata for downstream policy training.

\section{Method}
\label{sec:method}

\subsection{Downstream VLA Instantiation: GR00T N1.6}

To demonstrate the utility of SABER for downstream policy learning, we instantiate 
our study with \groot~\cite{groot_n1}, a modern multi-embodiment humanoid VLA. Our 
goal here is not to present a new backbone, but to evaluate how a strong existing 
policy stack responds to domain-specific retail data.

\nbf{VLM backbone}
A pretrained vision-language model encodes the robot's camera
observations together with the natural-language task instruction into a
joint embedding, providing strong visual-semantic grounding for parsing
product labels, spatial relationships, and task context.

\nbf{Flow-matching action head}
Rather than discretizing actions into tokens~\cite{rt2,openvla},
\groot uses a flow-matching formulation~\cite{pi0,groot_n1}.
The action head transforms a noise distribution into valid robot
trajectories conditioned on the VLM embedding, operating in continuous
joint-angle space for smooth, physically consistent motion.

\nbf{Pretraining distribution}
\groot was pretrained on large-scale teleoperated demonstrations spanning
single-arm manipulators, bimanual platforms, and full humanoid robots
across multiple datasets, including \texttt{singlepanda} and
\texttt{unified\_gr1}.
This gives the model strong priors for fundamental manipulation
primitives---reaching, grasping, placing, and articulated object
interaction---but entirely within laboratory and controlled simulation
settings.
The retail domain is absent.

\subsection{SABER Post-Training}

Modern VLA development increasingly follows a three-stage recipe: (1) internet-scale
vision-language pretraining, (2) large-scale robot action training on diverse 
manipulation datasets (e.g., Open X-Embodiment~\cite{openx}), and (3) domain- or
task-specific post-training. SABER is designed explicitly for stage~3. Rather than 
replacing broad robot pretraining, it supplies the missing deployment-domain 
signal: repeated retail skills, store-specific scene structure, and action 
distributions that are weakly represented in generic corpora. In our experiments,
we therefore start from pretrained \groot weights and use SABER as the dominant
domain-adaptation dataset. This preserves general manipulation priors while 
specializing the policy toward retail, without the prohibitive cost of retraining 
from scratch.

\begin{figure}[H]
  \centering
  \includegraphics[width=0.97\textwidth]{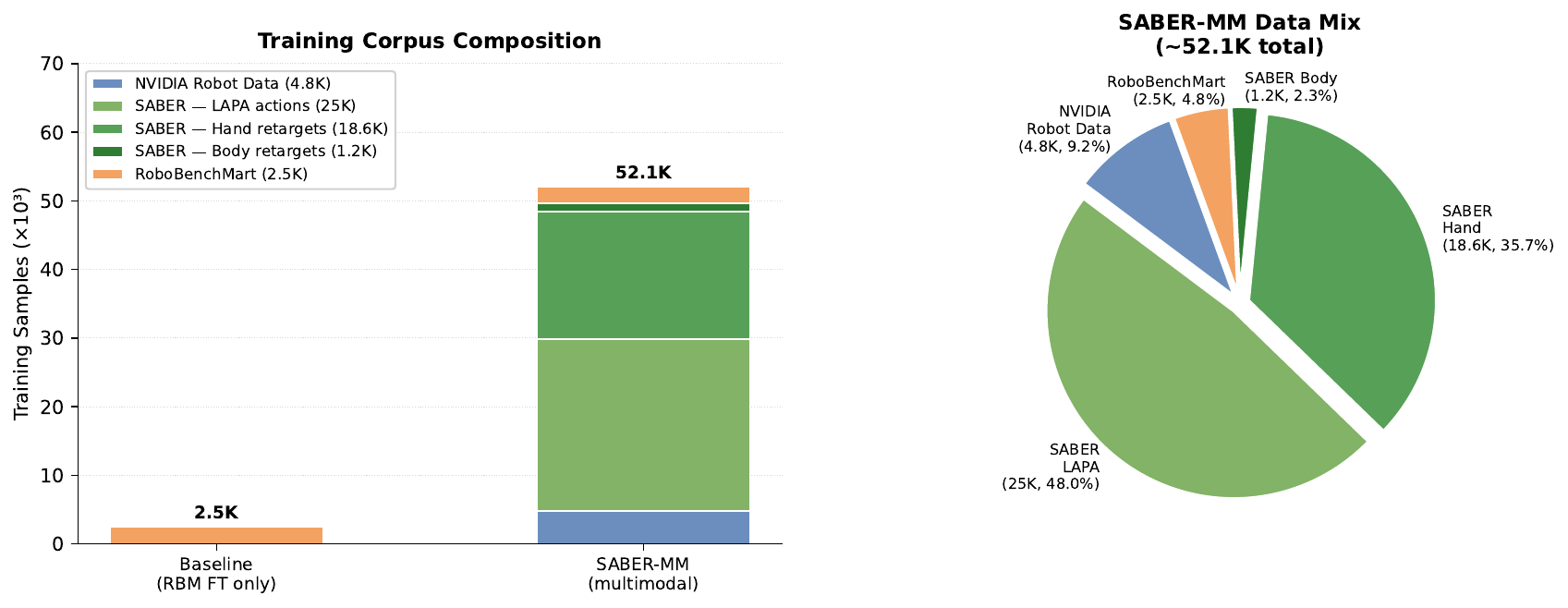}
  \caption{\textbf{SABER-MM training corpus composition.}
    \textit{Right}: pie chart of the SABER-MM mix ($\sim$52.1K total samples).
    SABER dataset components (85.9\%) comprise: latent actions from LAPA ($\sim$48.0\%), Dex-Retargeting hand
    retargets ($\sim$35.7\%), and body pose retargets ($\sim$2.3\%).
    The 4.8K robot-native anchor (\texttt{singlepanda},\texttt{unified\_gr1})
    provides stability during early training, and 2.5K RoboBenchMart task-aligned demonstrations are included.
    \textit{Note}: Percentages reflect dataset size. During training, batch-level weighting factors are applied to balance data distributions (see \cref{fig:batch_weighting}).}
  \label{fig:data_overview}
\end{figure}

\subsection{Post-Training Objective}
\label{sec:multitask}

We adopt GR00T N1's flow-matching objective \cite{hu2024adaflow, lipman2022flow,  liu2022flow, groot_n1} without modification. Each SABER stream - egocentric LAPA actions, egocentric retargeted dexterous-hand trajectories, ego-to-exo retargeted whole body robot trajectories are registered as  distinct embodiments with their own state encoder and action decoder, mirroring GR00T N1's native multi-embodiment design. The shared VLM backbone and diffusion transformer trunk are thus agnostic to stream-specific dimensionalities, allowing us to train on the standard conditional flow-matching loss
\begin{equation}
    \mathcal{L}_{\textit{fm}}(\theta) = \mathbb{E}_{\tau}\,[\lVert V_\theta(\phi_t, A^\tau_t, q_t) - (\epsilon - A_t) \bigr\rVert^2],
    \label{eq:flow_matching}
\end{equation}
where $A_t$ is a ground-truth action chunk, with flow matching timestep $\tau \in [0,1]$, and sampled noise $\epsilon \sim \mathcal{N}(0, I)$, $A^\tau_t = \tau\,A_t + (1-t)\,\epsilon$, model prediction $ V_\theta(\phi_t, A^\tau_t, q_t)$ with goal to approximate denoising vector field $ \epsilon - A_t$. No auxiliary losses or cross-stream consistency terms are introduced, so SABER serves as a drop-in extension of the GR00T N1 data mixture.

\subsection{Data Mixture Design}
\label{sec:mixture}

The training corpus combines three sources ($\sim$52.1K total): (1) \textbf{NVIDIA Robot Data}
(4.8K, 9.2\%) from \texttt{singlepanda} and \texttt{unified\_gr1}, providing essential
anchor signal for early stability, (2) \textbf{SABER} (44.8K, 85.9\%), the dominant signal,
comprising 25K LAPA tokens, 18.6K hand retargets, and 1.2K body pose retargets,
all ground-truth annotated, (3) \textbf{RoboBenchMart} (2.5K, 4.8\%), task-aligned
sim demonstrations.

\begin{figure}[H]
  \centering
  \includegraphics[width=0.95\textwidth]{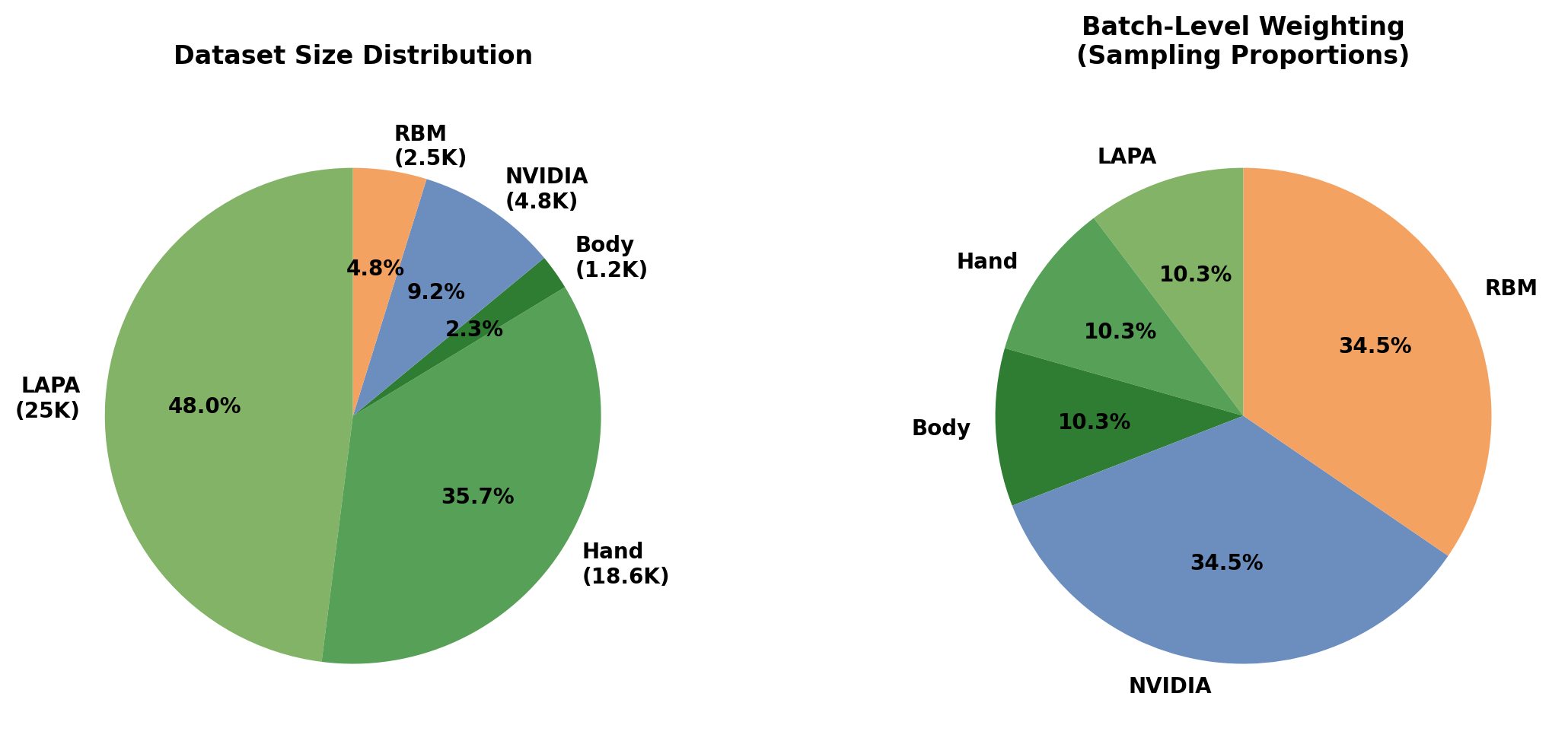}
  \caption{\textbf{Dataset size vs.\ batch-level weighting.}
    \textit{Left}: Distribution of data by actual dataset size (dataset composition).
    \textit{Right}: Distribution during training via batch-level weighting factors.
    While SABER dominates by size (85.9\%), the weighting strategy ensures that each batch has balanced exposure to retail-specific data (SABER, 30.9\%), general robot manipulation priors (NVIDIA, 34.55\%), and task-aligned simulation (RoboBenchMart, 34.55\%). This prevents the larger SABER corpus from overwhelming historical priors and ensures robust generalization across all three data distributions.}
  \label{fig:batch_weighting}
\end{figure}

\nbf{Batch-level data weighting}
Although the datasets vary substantially in size, during training we apply explicit weighting factors when sampling each batch (the size of a batch during post-training is 896). Rather than sampling proportionally to dataset size, we use: LAPA 10.3\%, Hand-retargets 10.3\%, Body-retargets 10.3\%, NVIDIA 34.55\%, and RoboBenchMart 34.55\%. This weighting strategy serves two purposes. First, it normalizes the influence of our retail-specific SABER streams (30.9\% combined) against the historical NVIDIA priors (34.55\%), preventing the larger SABER corpus from overwhelming the model's general manipulation skills. Second, it ensures that each batch has balanced exposure to the three distinct data distributions: retail-specific capture (SABER), general robot manipulation (NVIDIA), and the evaluation task domain (RoboBenchMart). This balanced sampling improves generalization by preventing the model from overfitting to any single data source's statistical properties.

\section{Experiments}
\label{sec:experiments}
We evaluate SABER on a standardized retail manipulation benchmark, measuring the impact of domain-specific post-training against simulation-only and label-free baselines. 

\subsection{Evaluation Benchmark}
We evaluate on ten tasks from \robobenchmart~\cite{robobenchmart},
covering four categories and Fetch(mobile robot + panda arm ) robot is used for all the tasks :\\
\textbf{Fridge} (2 tasks): \texttt{open\_fridge}, \texttt{close\_fridge}- open/close a sliding door of the fridge.\\
\textbf{Move from board to board} (3 tasks): \texttt{duff}, \texttt{nestle}, \texttt{vanish} - Move the target item from one shelf to another.\\
\textbf{Pick from floor} (2 tasks): \texttt{beans}, \texttt{slam} - Pick the target item from the floor and place in on the shelf.\\
\textbf{Pick to basket} (3 tasks): \texttt{fanta}, \texttt{nivea}, \texttt{stars} - Pick the target item from shelf and place it in the basket. 

For our evaluation we perform 100 roll-outs for each task and success is decided if the rollout has reached this condition by the end of the rollout: "robot joint velocities < 0.2" and "target object reached target position" and "Non target objects were not disturbed". Anything else is considered a failure.

\nbf{Task Progress Score (TPS)}
Binary success metrics, while useful, fail to capture task progress information - a robot may grasp an object but fail to place it at the target, losing a valuable intermediate signal. To address this, we further divide each task into four ordinal stages $\{0,\,\tfrac{1}{3},\,\tfrac{2}{3},\,1\}$: (0) failure or scene disturbance; (1/3) robot approaches target without disturbance; (2/3) robot grasps target object without disturbance; (1) full success. Fridge tasks are structurally simpler and do not subdivide meaningfully. We apply TPS to all non-fridge tasks, which are all variants of pick-and-place manipulation with consistent milestones across categories. 

\begin{itemize}[leftmargin=*,nosep]
  \item \textbf{Board to Board}: (1) robot approaches target product;
        (2) successful grasp; (3) product placed on destination board.
  \item \textbf{Floor Pick}: (1) robot lowers toward item;
        (2) successful grasp at floor level; (3) item retrieved to standing height.
  \item \textbf{Basket Pick}: (1) robot approaches target item;
        (2) successful grasp; (3) item placed into basket.
\end{itemize}

\subsection{Experimental Configuration}

\nbf{SABER-MM (multimodal post-training)}
The downstream VLA is post-trained on the full three-stream SABER corpus:
4.8K NVIDIA Robot Data (anchor) + 25K LAPA latent actions + 18.6K hand pose retargets + 1.2K body pose retargets + 2.5K RoboBenchMart task-aligned demonstrations
($\sim$52.1K total samples).
This configuration combines all three action representation streams and is evaluated at the empirically optimal checkpoint of 150K iterations. 

\section{Results}
\label{sec:results}

\subsection{Dataset and Annotation Quality}

SABER comprises $\sim$44.8K training samples derived from 100+ hours of natural retail activity,
annotated through a rigorous multi-stage QC process. This
ground-truth annotated corpus, combined with embodiment-agnostic retargeting, is the primary
contribution and can be applied to any downstream VLA or robotics pipeline.

\subsection{Model Post-Training Results}

\begin{table}[t]
\centering
\caption{\textbf{Overall success rates: Baseline vs.\ SABER-MM.}
  SABER-MM evaluation on ten \robobenchmart tasks compared to baseline GR00T-N1.6 with RoboBenchMart fine-tuning only.}
\label{tab:ablation_success}
\small
\begin{tabular}{l|cc}
\toprule
\textbf{Task} & \textbf{Baseline (RBM FT)} & \textbf{SABER-MM} \\
\midrule
\texttt{close\_fridge} & 0.74 & 1.00 \\
\texttt{open\_fridge} & 0.12 & 0.82 \\
\midrule
\texttt{move\_from\_board\_to\_board\_duff} & 0.10 & 0.1 \\
\texttt{move\_from\_board\_to\_board\_nestle} & 0.02 & 0.02 \\
\texttt{move\_from\_board\_to\_board\_vanish} & 0.02 & 0.11 \\
\midrule
\texttt{pick\_from\_floor\_beans} & 0.04 & 0.17 \\
\texttt{pick\_from\_floor\_slam} & 0.02 & 0.17 \\
\midrule
\texttt{pick\_to\_basket\_fanta} & 0.08 & 0.19 \\
\texttt{pick\_to\_basket\_nivea} & 0.08 & 0.21 \\
\texttt{pick\_to\_basket\_stars} & 0.12 & 0.14 \\
\midrule
\textbf{Mean (all tasks)} & 0.134 & 0.293 \\
\textbf{Mean (non-fridge)} & 0.060 & 0.138 \\
\bottomrule
\end{tabular}
\end{table}

\cref{tab:ablation_success} presents overall success rates for SABER-MM post-training on ten \robobenchmart tasks, compared to the baseline GR00T-N1.6 model with RoboBenchMart fine-tuning only. The success rates from SABER-MM post-training demonstrate substantial gains: 29.3\% mean success versus 13.4\% baseline—a \textbf{2.19x improvement}. Fridge tasks (refrigerator interaction) achieve particularly strong results with SABER-MM, reaching 100\% and 82\% success on close and open operations respectively, while non-fridge tasks (board-to-board, floor, basket) show consistent improvement with 13.88\% mean success versus 6.0\% baseline. \cref{tab:main_results} reports per-task Task Progress Scores for SABER-MM post-training over non-fridge tasks, showing that improvements extend across partial task milestones, not merely full success rates. The Task Progress Score (P$\geq\!1/3$, P$\geq\!2/3$, P$=1$) reveals that models consistently achieve partial sub-goal completion even on tasks where full success remains rare, indicating systematic progress through manipulation sequences.

\nbf{Task Progress Score interpretation}
SABER-MM post-training demonstrates genuine task progression across retail manipulation. On non-fridge tasks, SABER-MM achieves mean P$\geq\!1/3$ of 0.74375 (task initiation), mean P$\geq\!2/3$ of 0.445 (substantial progress), and mean P$=1$ of 0.14625 (full success)—substantially outperforming baseline means of 0.838, 0.278, and 0.066 respectively. This improvement indicates that SABER-MM teaches models to progress further through each task sequence, not merely initiating tasks more frequently. The gap between P$\geq\!1/3$ and P$=1$ remains pronounced on board and basket tasks, indicating that reaching and grasping are well-learned, but precise placement and final positioning remain challenging---a distinction that binary success rates alone would obscure. These results suggest that high-fidelity multimodal action supervision provides a meaningful signal for downstream retail policy learning.

\begin{table}[H]
\centering
\caption{\textbf{Task Progress Score across \robobenchmart tasks: Baseline vs.\ SABER-MM.}
  Three progress levels shown: P$\geq\!1/3$ (initiated), P$\geq\!2/3$ (substantial), P$=1$ (full success).
  SABER-MM enables progression through task milestones compared to baseline GR00T-N1.6.
  Note: Small differences between Task Progress Score P$=1$ (full success) and binary success rates arise from inherent stochasticity in policy evaluation runs - separate evaluation batches can yield slightly different success rates due to random seeds and environment variations.}
\label{tab:main_results}
\small
\begin{tabular}{l|ccc|ccc}
\toprule
\textbf{Task} & \multicolumn{3}{c|}{\textbf{Baseline}} & \multicolumn{3}{c}{\textbf{SABER-MM}} \\
& $\geq\!1/3$ & $\geq\!2/3$ & $=1$ & $\geq\!1/3$ & $\geq\!2/3$ & $=1$ \\
\midrule
\texttt{move\_from\_board\_to\_board\_duff} & 0.87 & 0.51 & 0.13 & 0.85 & 0.54 & 0.11 \\
\texttt{move\_from\_board\_to\_board\_nestle} & 0.96 & 0.12 & 0.01 & 0.73 & 0.27 & 0.02 \\
\texttt{move\_from\_board\_to\_board\_vanish} & 0.98 & 0.24 & 0.02 & 0.91 & 0.60 & 0.11 \\
\midrule
\texttt{pick\_from\_floor\_beans} & 0.86 & 0.48 & 0.1 & 0.83 & 0.65 & 0.20 \\
\texttt{pick\_from\_floor\_slam} & 1.00 & 0.30 & 0.05 & 0.91 & 0.64 & 0.18 \\
\midrule
\texttt{pick\_to\_basket\_fanta} & 0.60 & 0.21 & 0.05 & 0.52 & 0.29 & 0.19 \\
\texttt{pick\_to\_basket\_nivea} & 0.81 & 0.21 & 0.08 & 0.69 & 0.31 & 0.21 \\
\texttt{pick\_to\_basket\_stars} & 0.63 & 0.15 & 0.09 & 0.51 & 0.26 & 0.15 \\
\midrule
\textbf{Mean} & 0.838 & 0.278 & 0.066 & 0.74375 & 0.445 & 0.14625 \\
\bottomrule
\end{tabular}
\end{table}
\subsection{Qualitative Rollout Analysis}

\begin{figure}[t]
  \centering
  \includegraphics[width=0.97\textwidth]{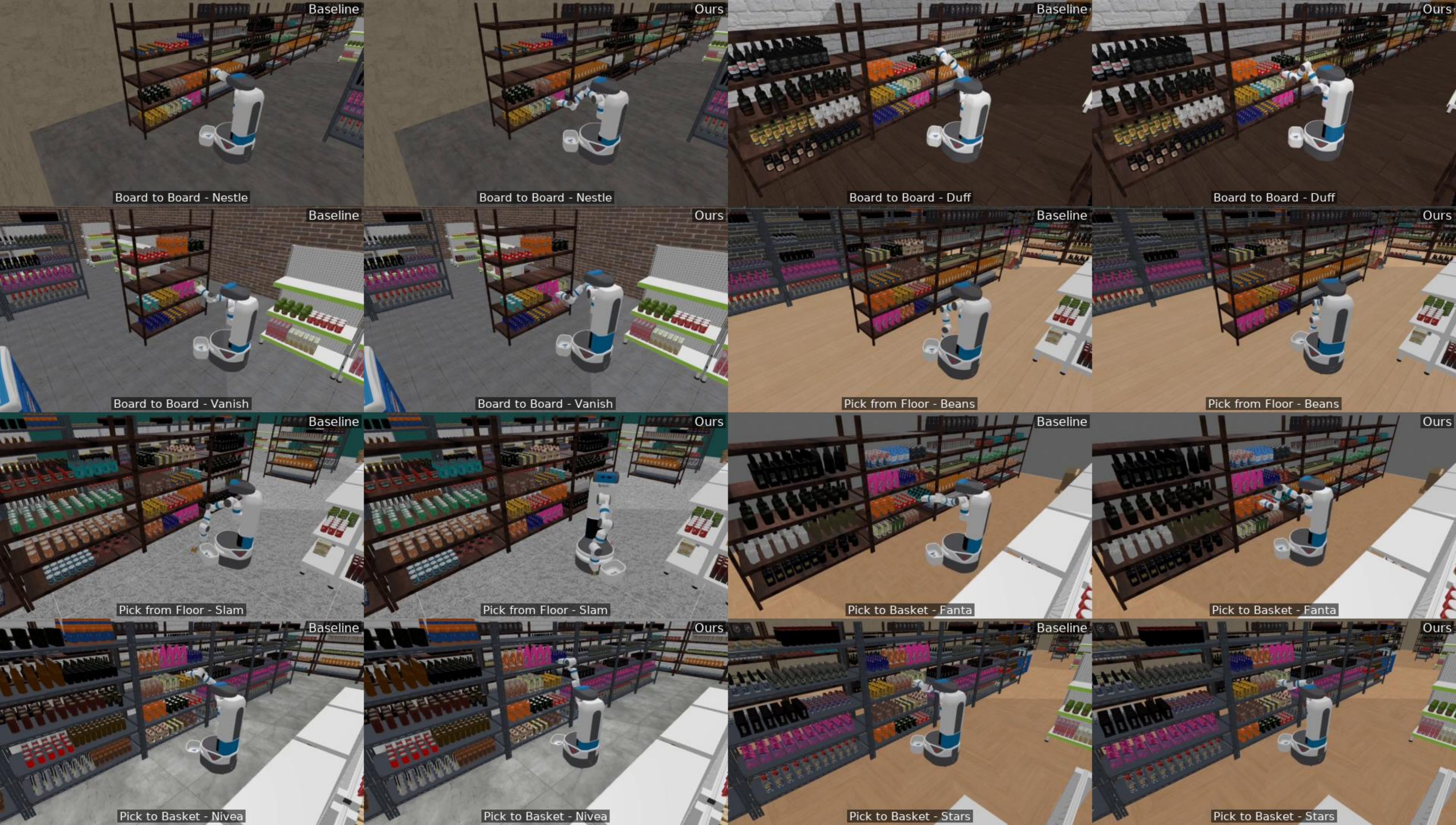}
  \caption{\textbf{SABER post-training qualitative results on \robobenchmart.}
    Each panel pair shows \textit{Baseline} (base \groot without post-training)
    versus \textit{Ours} (SABER post-trained) on eight representative
    \robobenchmart tasks spanning all four categories:
    board-to-board (Nestl\'{e}, Duff, Vanish), floor pick (Beans, Slam),
    and basket pick (Fanta, Stars).
    SABER post-training produces markedly more purposeful approach
    trajectories and better object contact across all categories.}
  \label{fig:rbm_collage}
\end{figure}

\cref{fig:rbm_collage} contrasts rollouts of the base \groot model
(Baseline) against our SABER post-trained checkpoint across eight
representative \robobenchmart tasks.
Across board-to-board, floor-pick, and basket-pick categories, SABER
post-training produces markedly more purposeful approach trajectories
and higher-quality object contact events.
The improvement is most pronounced on tasks requiring precise hand shaping
(Vanish spray, Fanta can) where the Dex-Retargeting supervision provides
explicit finger-joint priors that the base model lacks.

Successful fridge rollouts exhibit smooth, purposeful trajectories: the
robot approaches the handle along the correct axis, makes firm contact,
and completes the articulation in a single continuous motion---the LAPA
latent tokens from human fridge-opening clips have transferred a genuine
kinematic prior.
The Dex-Retargeting stream is expected to produce the most visible improvement
on board and basket tasks, where precise gripper finger shaping around
small product packaging is the dominant failure mode.
The body pose stream is expected to extend floor-pick success rates by
providing explicit whole-body reach coordination---the robot's downward
extension is currently the primary failure on floor tasks.

\begin{figure}[H]
    \centering
    \setlength{\tabcolsep}{2pt}%
    \renewcommand{\arraystretch}{1.1}%
    \begin{tabular}{@{}c@{\hspace{4pt}}ccc@{}}

        \multicolumn{4}{@{}l}{\small\textbf{Task 1: Move Duff Beer from shelf to shelf}} \\[2pt]
        \rotatebox{90}{\small\textbf{(a) Fail}} &
        \includegraphics[width=0.32\linewidth]{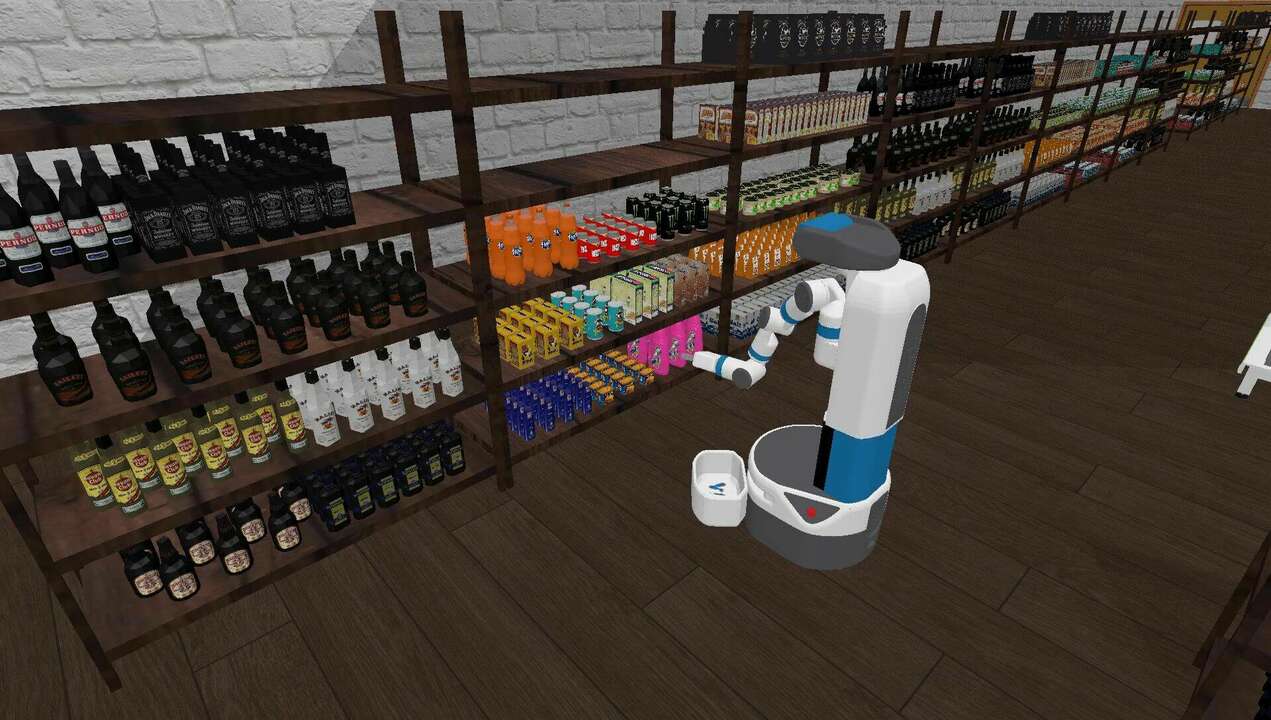}  &
        \includegraphics[width=0.32\linewidth]{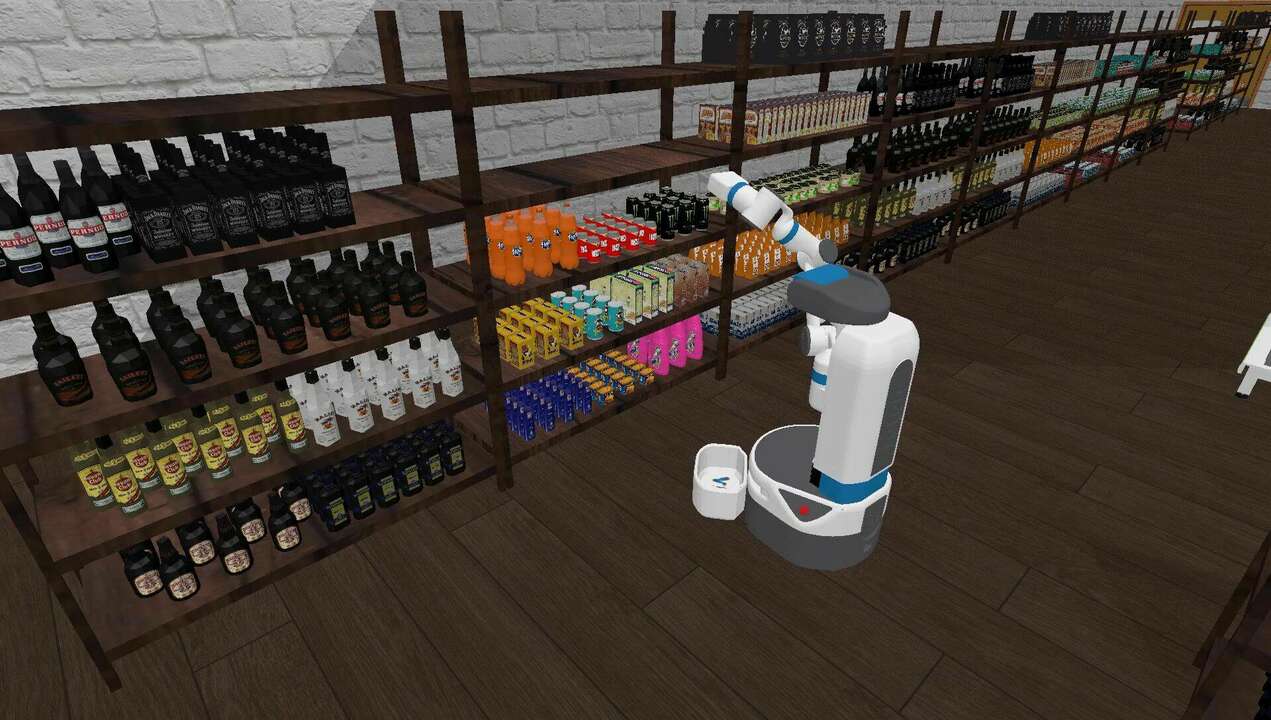}  &
        \includegraphics[width=0.32\linewidth]{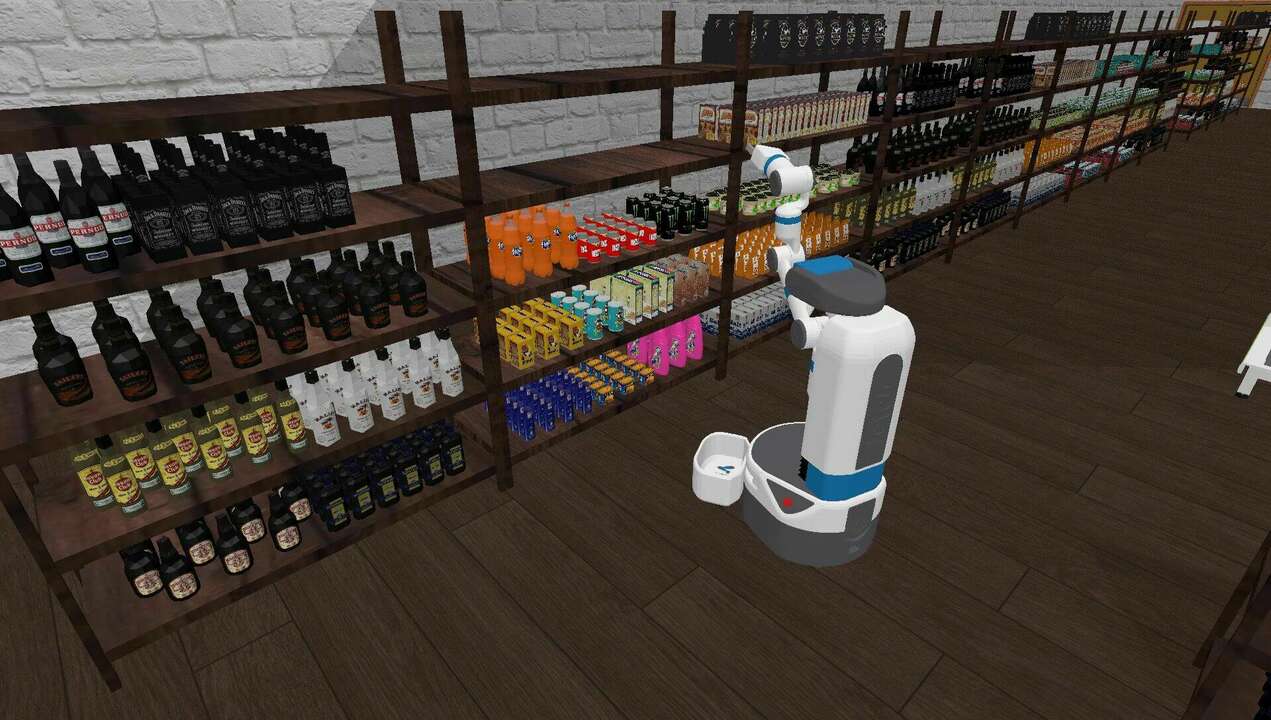} \\
        \rotatebox{90}{\small\textbf{(b) Pass}} &
        \includegraphics[width=0.32\linewidth]{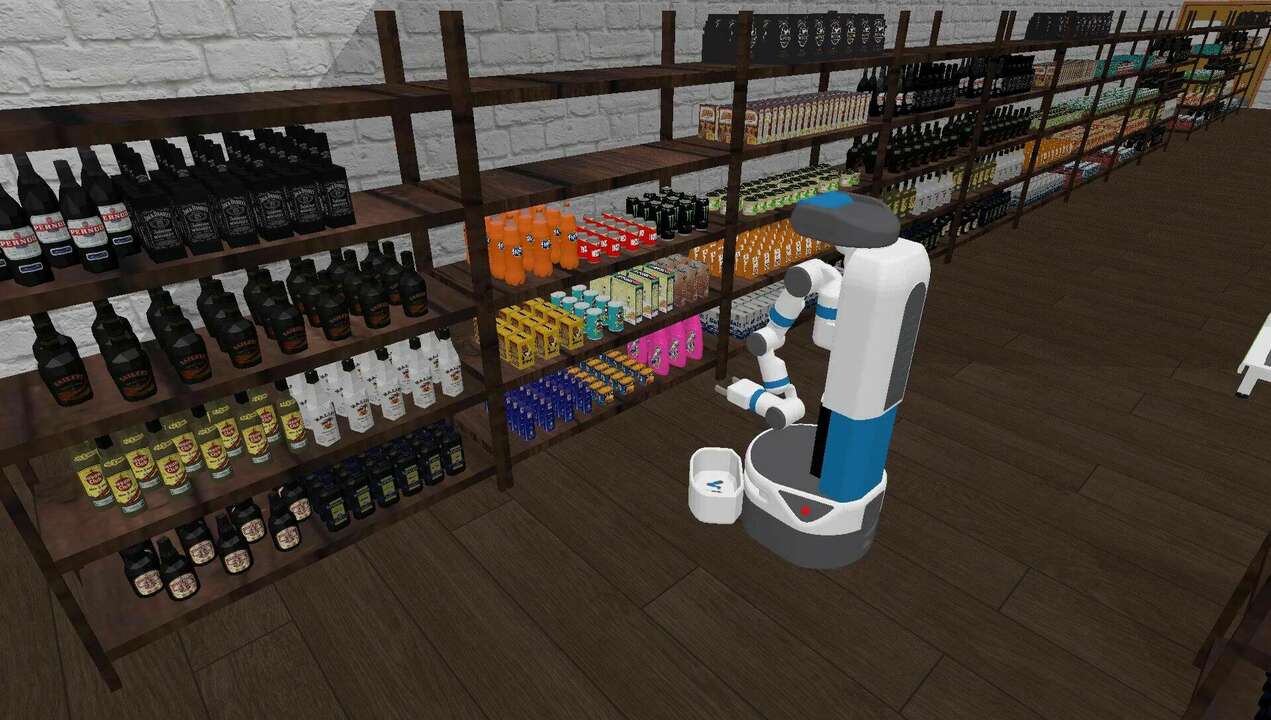}  &
        \includegraphics[width=0.32\linewidth]{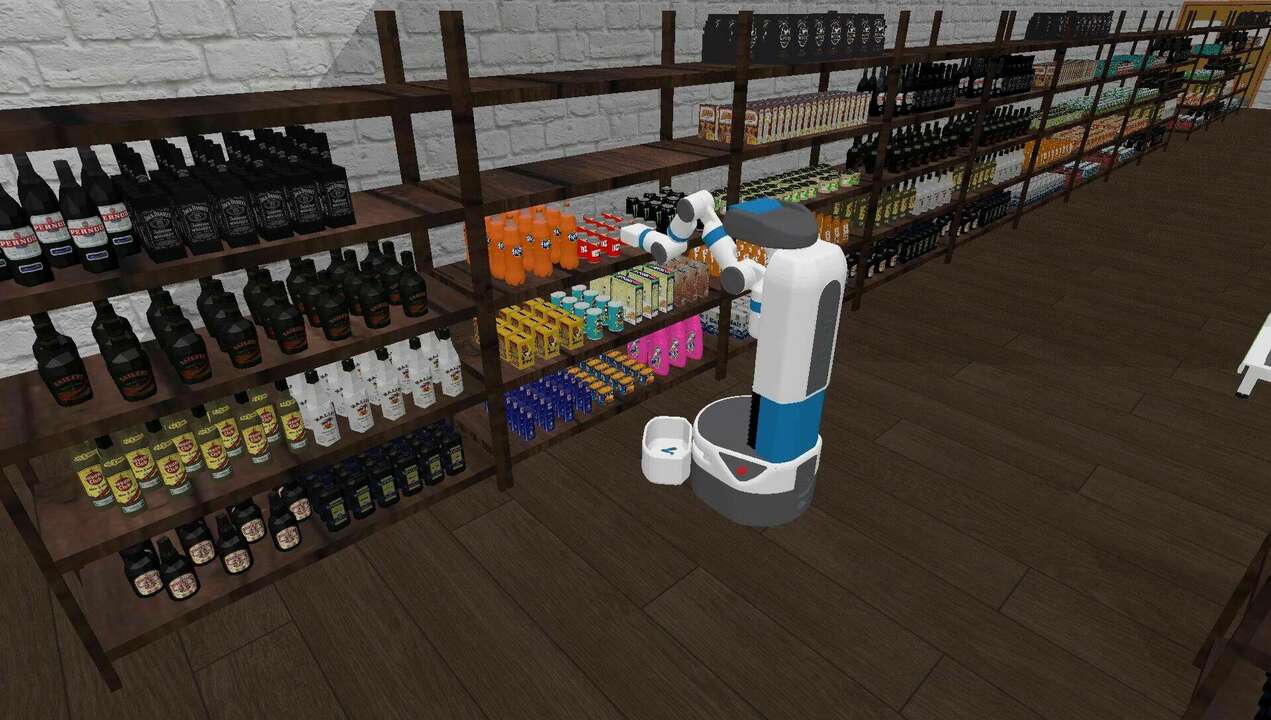}  &
        \includegraphics[width=0.32\linewidth]{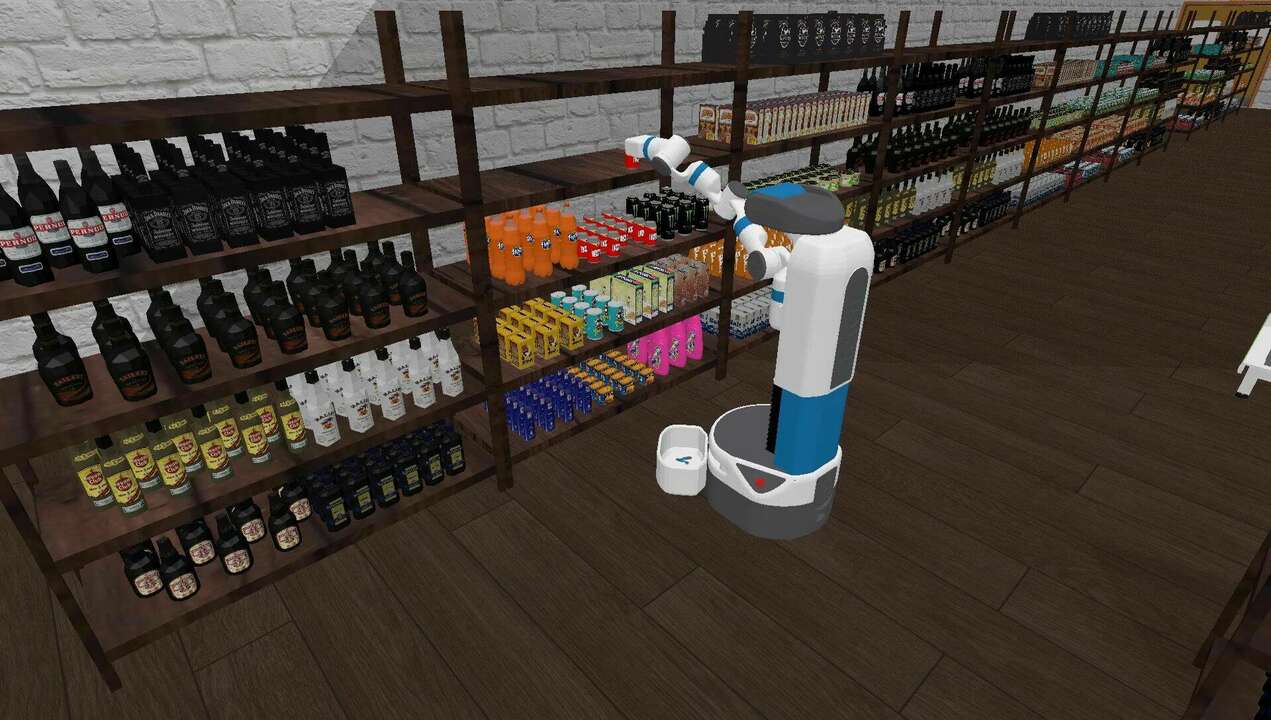} \\
         & \small $t{=}3\,$s & \small $t{=}8\,$s & \small $t{=}17\,$s \\[6pt]

        \multicolumn{4}{@{}l}{\small\textbf{Task 2: Pick Beans from floor}} \\[2pt]
        \rotatebox{90}{\small\textbf{(a) Fail}} &
        \includegraphics[width=0.32\linewidth]{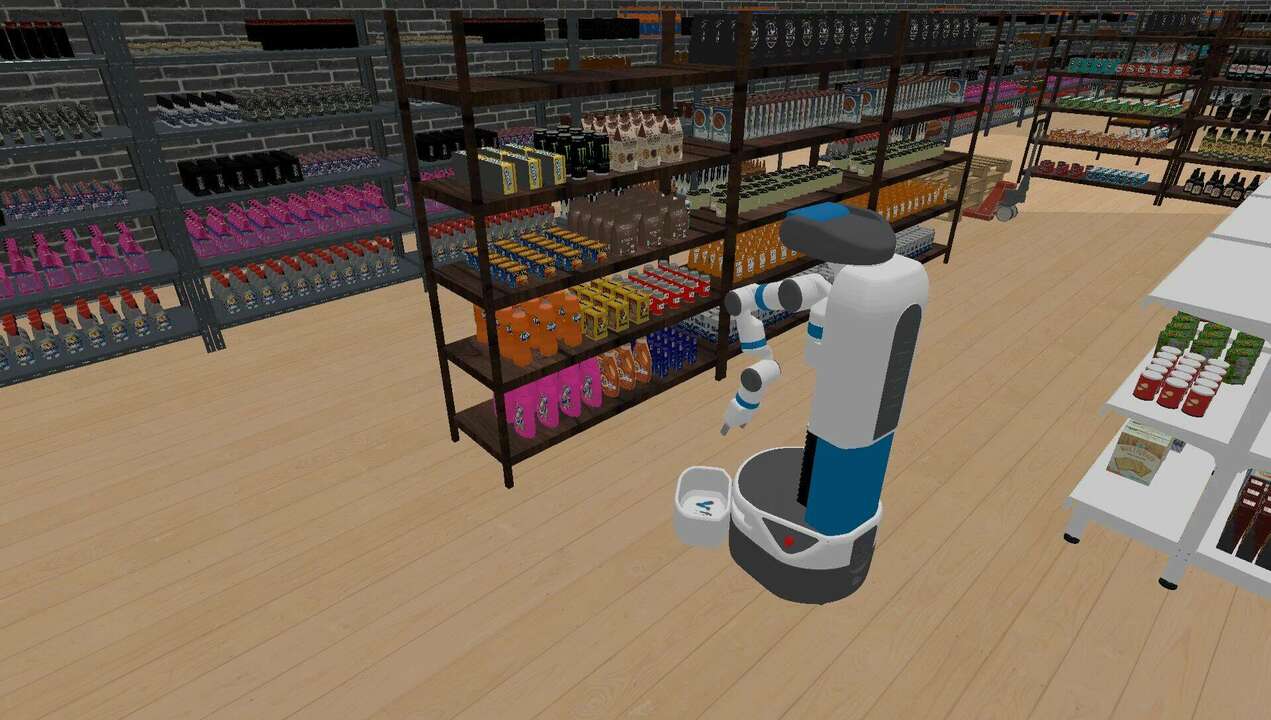}  &
        \includegraphics[width=0.32\linewidth]{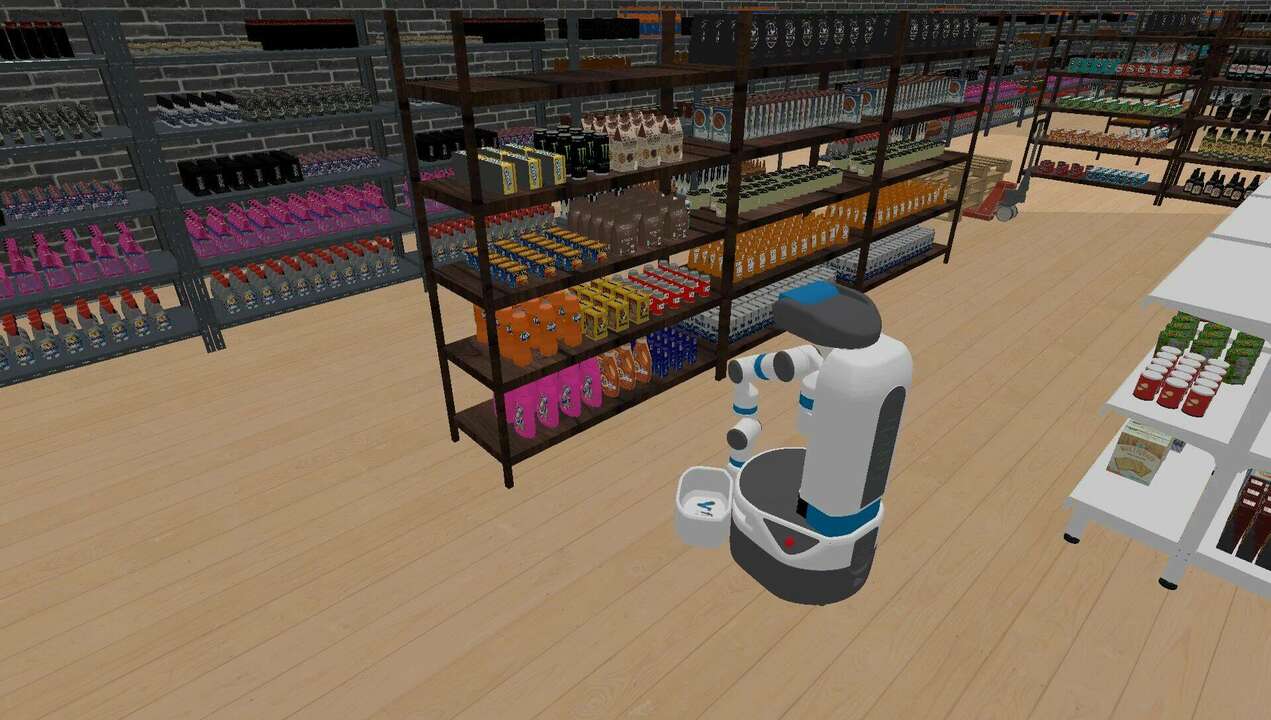}  &
        \includegraphics[width=0.32\linewidth]{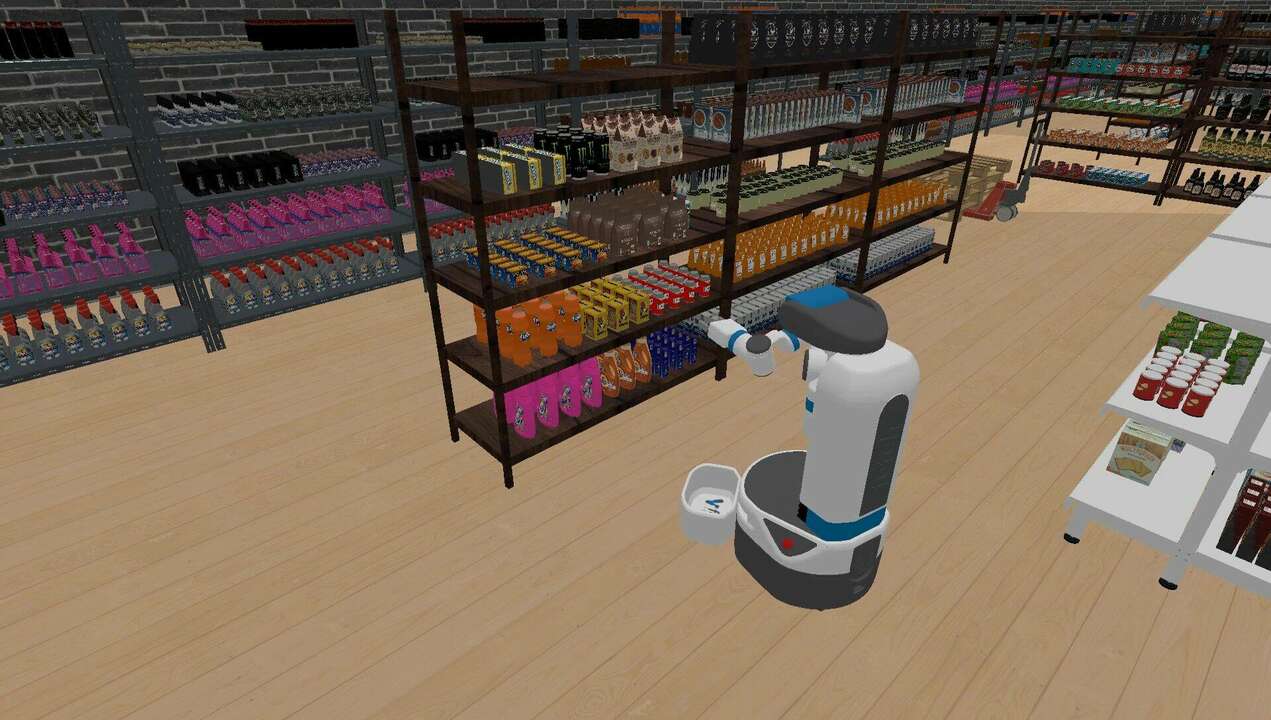} \\
        \rotatebox{90}{\small\textbf{(b) Pass}} &
        \includegraphics[width=0.32\linewidth]{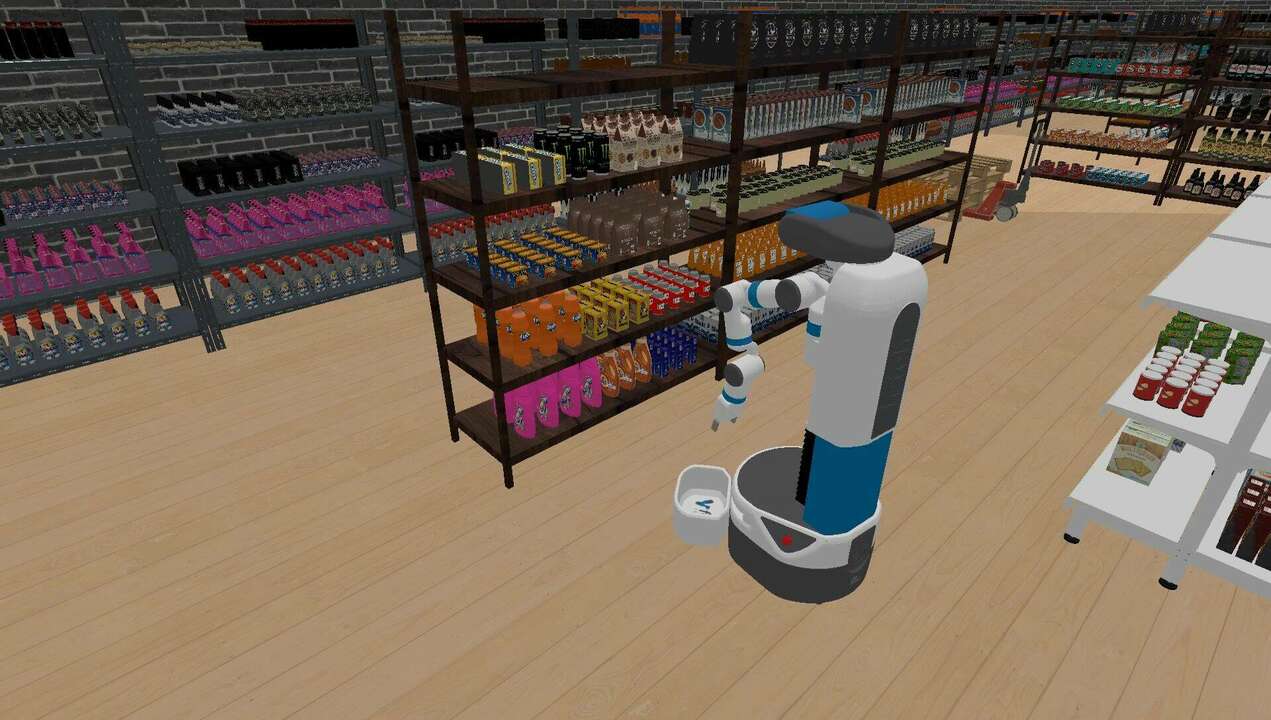}  &
        \includegraphics[width=0.32\linewidth]{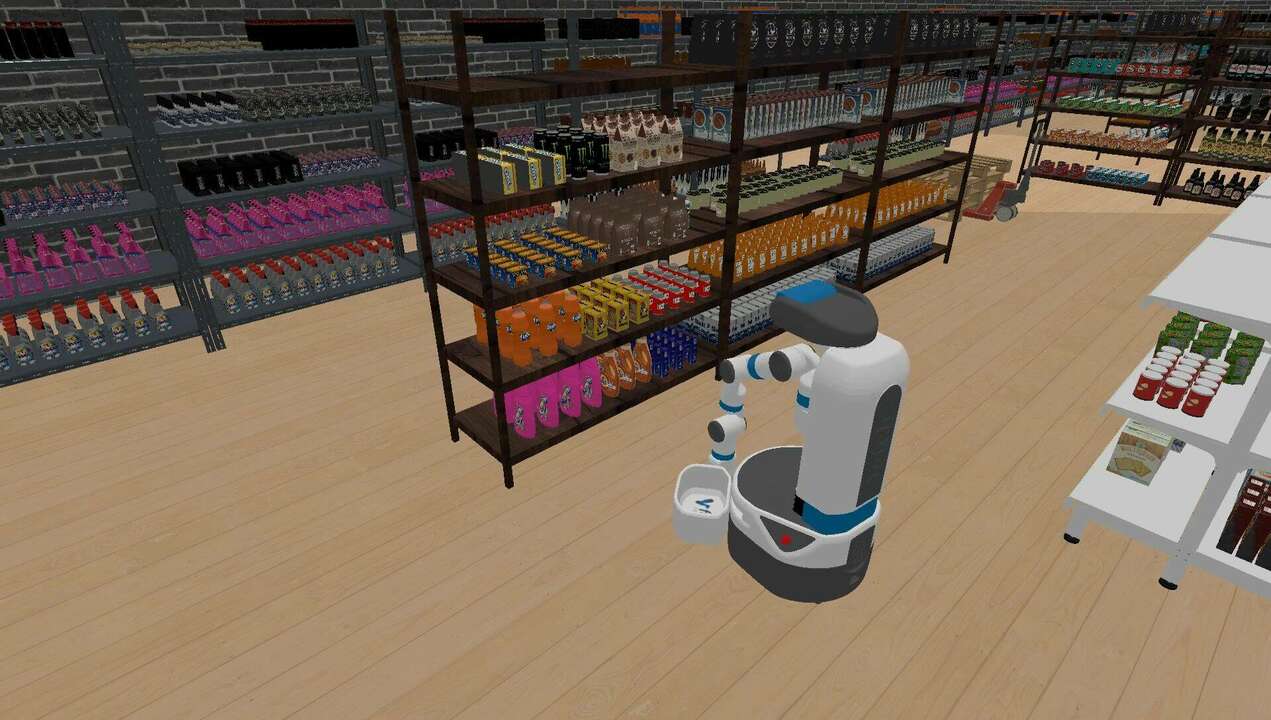}  &
        \includegraphics[width=0.32\linewidth]{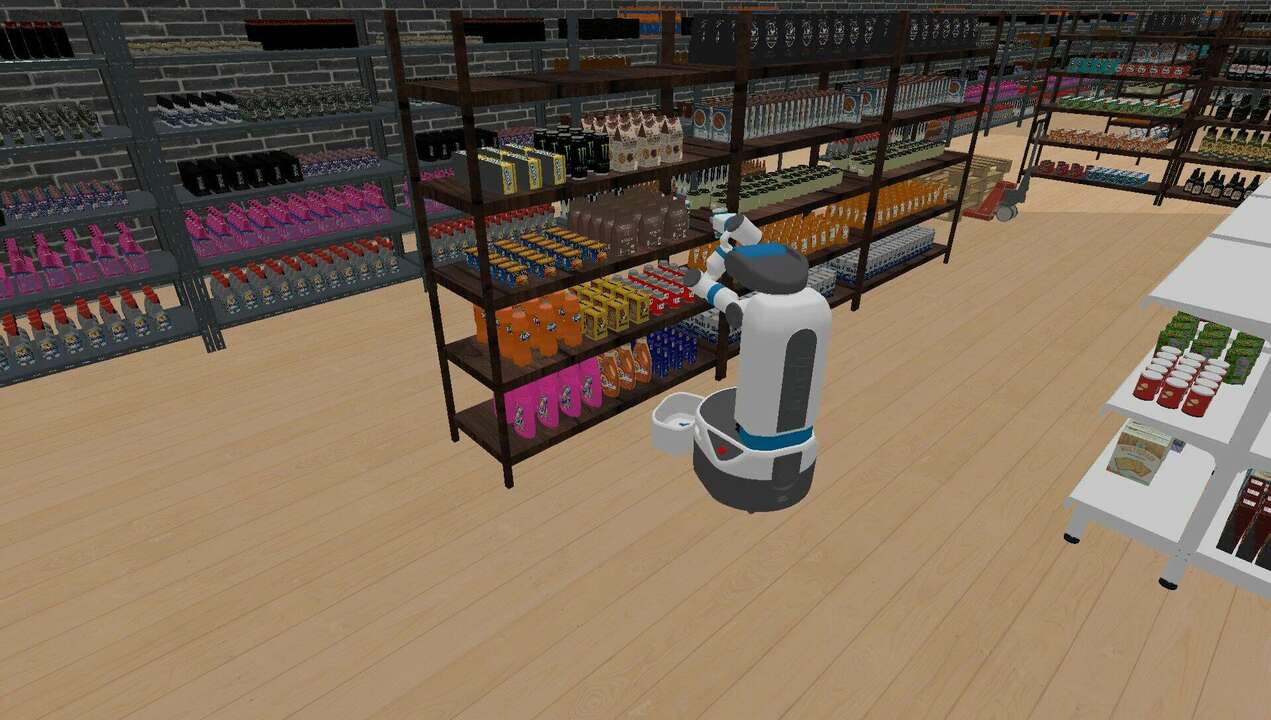} \\
         & \small $t{=}3\,$s & \small $t{=}5\,$s & \small $t{=}14\,$s \\[6pt]

        \multicolumn{4}{@{}l}{\small\textbf{Task 3: Pick Fanta to basket}} \\[2pt]
        \rotatebox{90}{\small\textbf{(a) Fail}} &
        \includegraphics[width=0.32\linewidth]{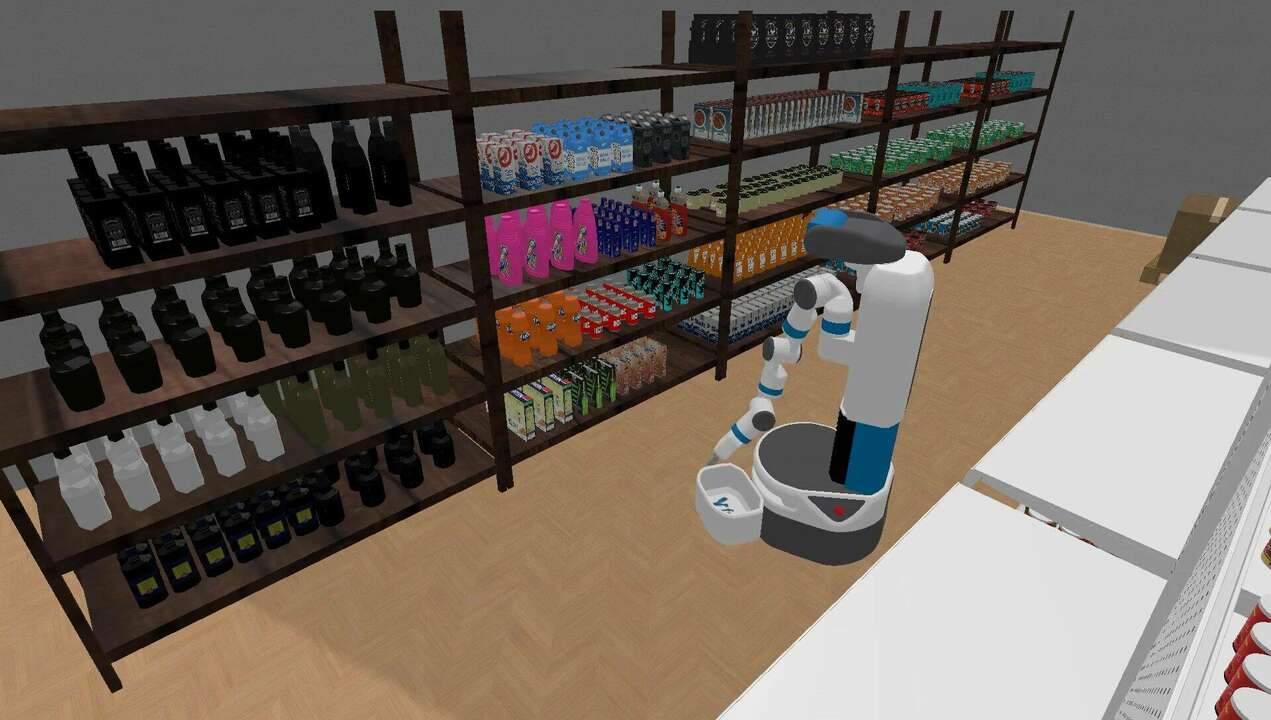}  &
        \includegraphics[width=0.32\linewidth]{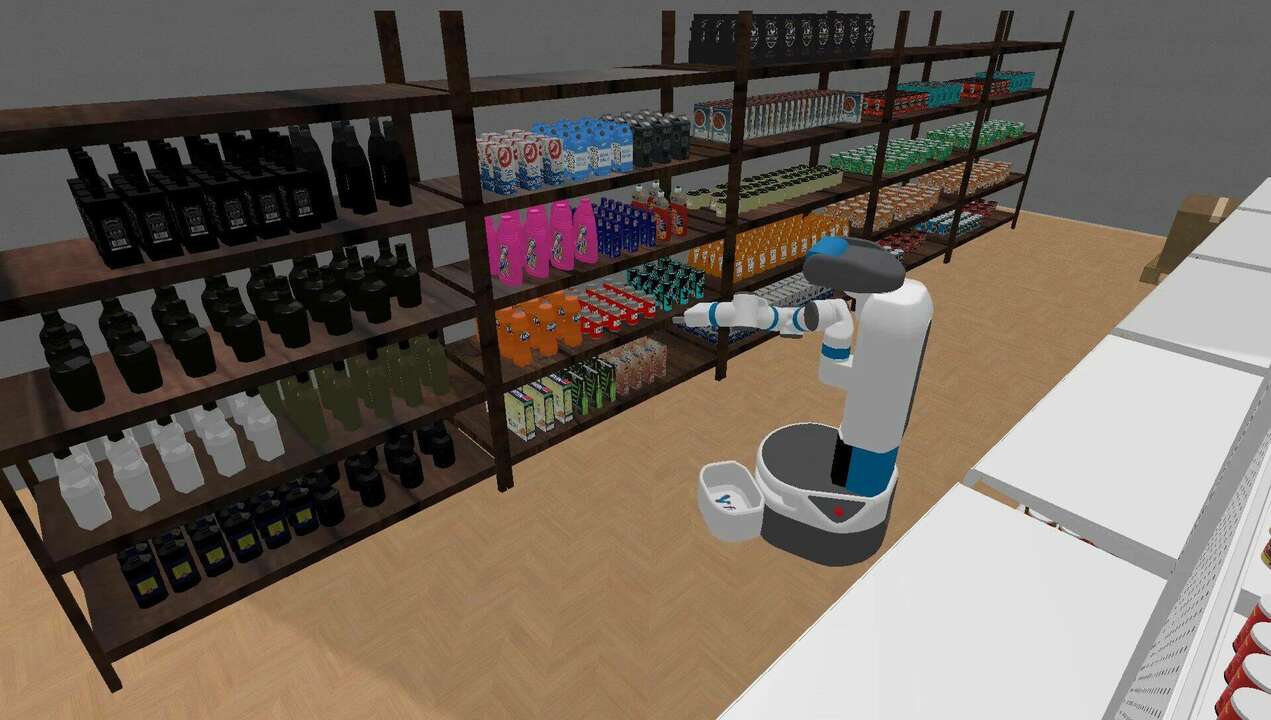}  &
        \includegraphics[width=0.32\linewidth]{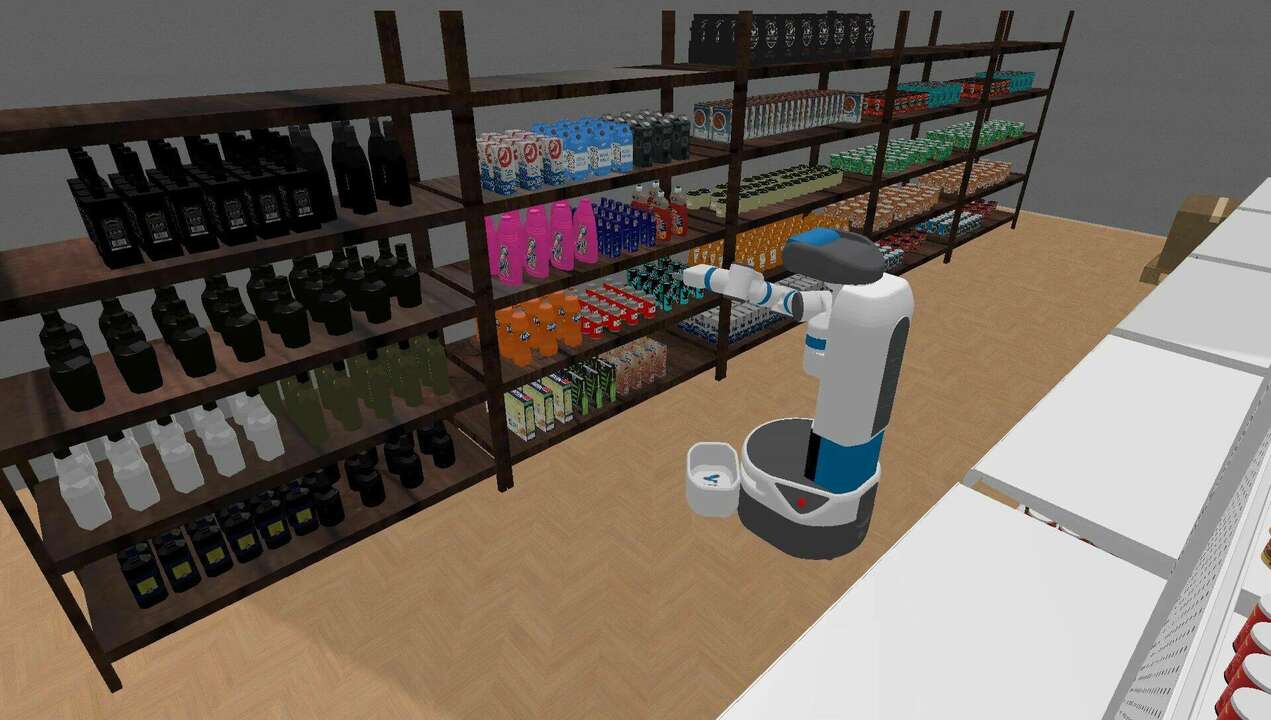} \\
        \rotatebox{90}{\small\textbf{(b) Pass}} &
        \includegraphics[width=0.32\linewidth]{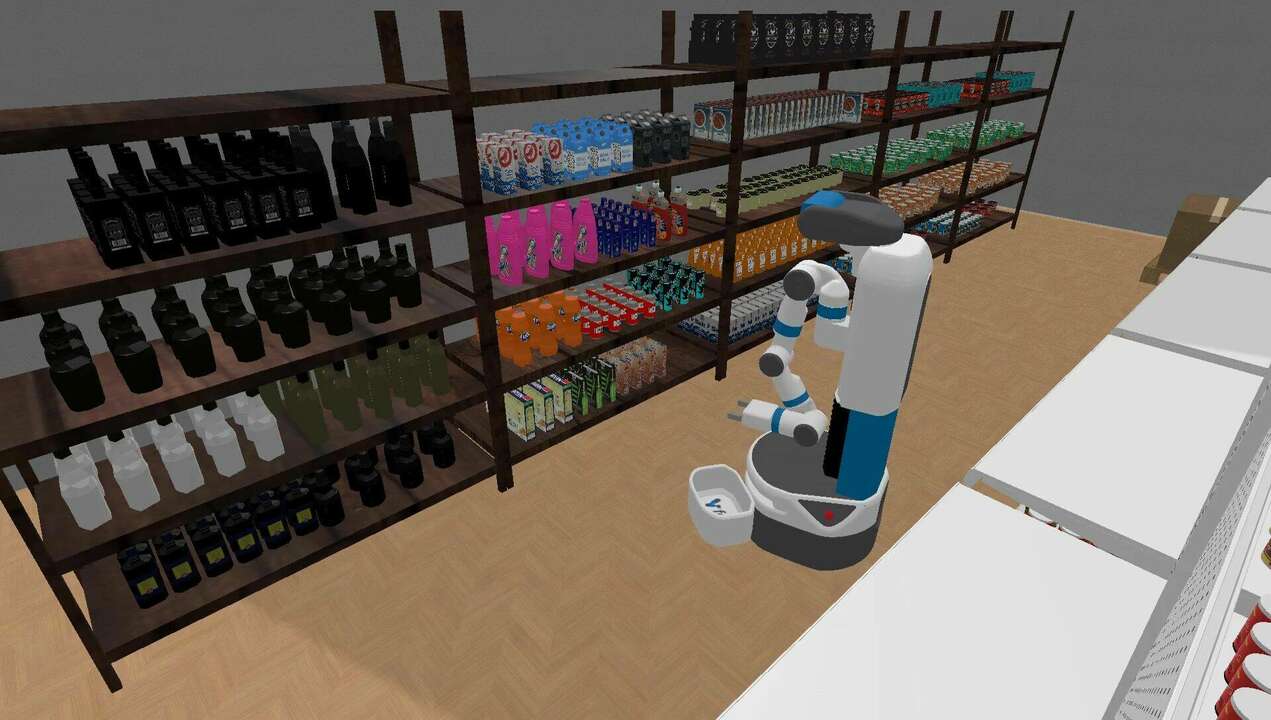}  &
        \includegraphics[width=0.32\linewidth]{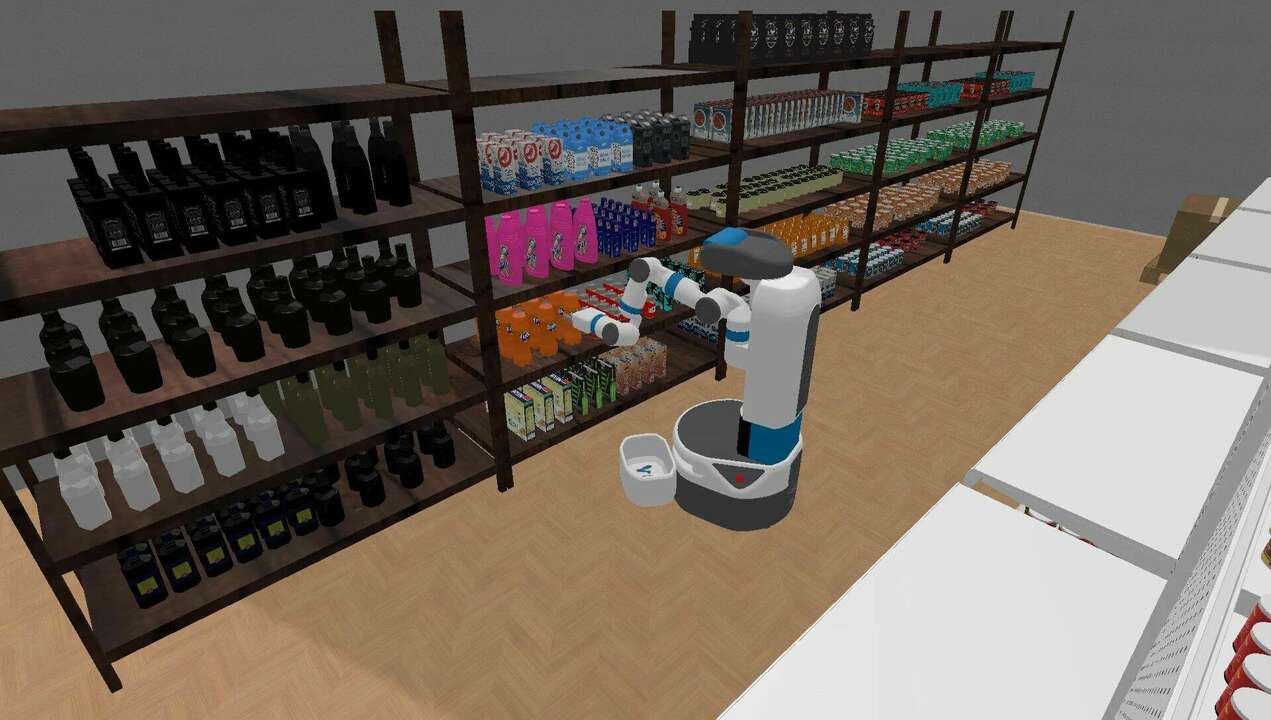}  &
        \includegraphics[width=0.32\linewidth]{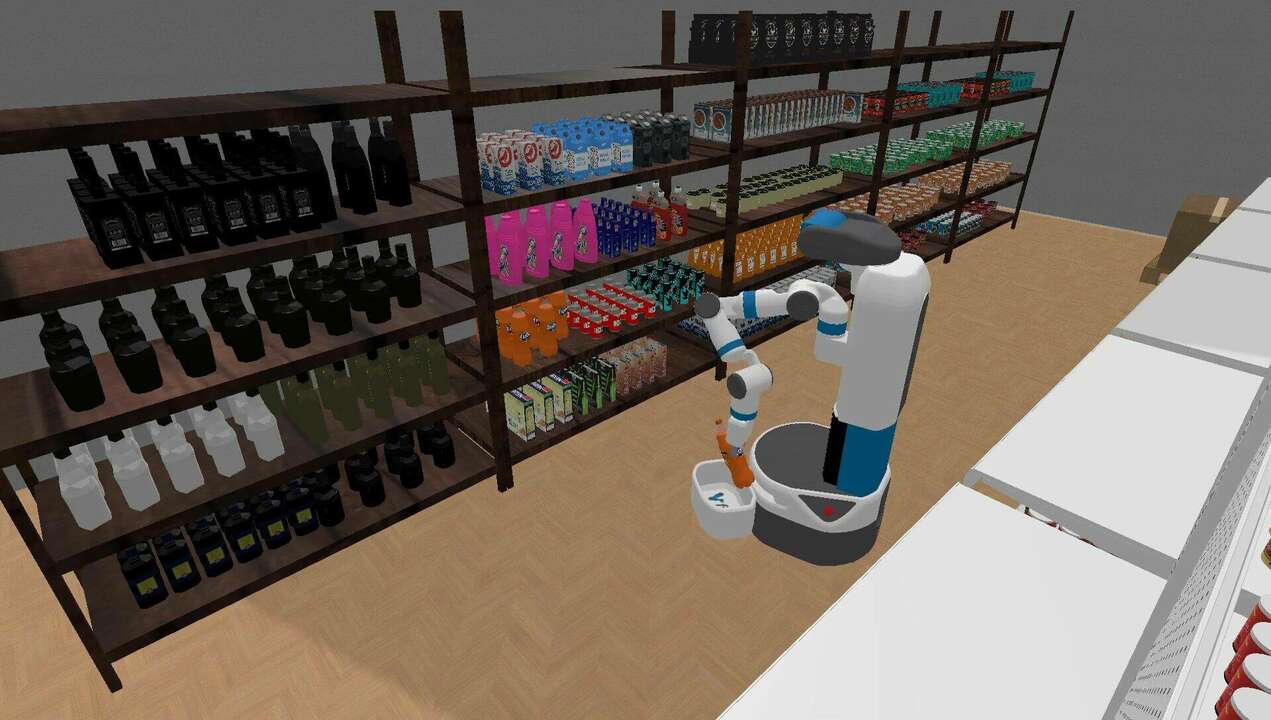} \\
         & \small $t{=}3\,$s & \small $t{=}8\,$s & \small $t{=}11\,$s \\[6pt]

    \end{tabular}
    \caption{Qualitative comparison of policy rollouts across multiple retail manipulation tasks.
    For each task, the top row \textbf{(a)} shows a failure rollout and the bottom row \textbf{(b)} shows a successful rollout, sampled at three timestamps along the trajectory.}
    \label{fig:pass_fail_qualitative}
\end{figure}

Building on this, \cref{fig:pass_fail_qualitative} presents three frames sampled along the trajectory of a failure rollout and a successful rollout for each task, allowing a direct visual comparison of where the two diverge. In each task block, row (a) (``Fail'') shows the rollout produced by the baseline policy, while row (b) (``Pass'') shows the rollout produced by our post-trained model. The rollouts from the post-trained model show better completion and progress than the baseline.

\section{Discussion}
\label{sec:discussion}

\nbf{Complementarity of three representation streams}
The core hypothesis behind SABER's multi-stream design is that LAPA,
hand, and body retargets provide non-overlapping kinematic
information.
LAPA tokens are strong on whole-arm trajectory and visually recognizable
contact events, but weaker on finger articulation and full-body
coordination.
Dex-Retargeting provide finger-level precision that LAPA cannot
encode---the explicit joint angles of a pinch grasp or precision grip,
retargeted to robot space.
Body pose retargets supply the only source of torso-arm-leg coordination
in the corpus, essential for floor retrieval and extended-reach tasks.
Training all three streams jointly allows the shared backbone to build
representations that are informed by all three levels of kinematic
abstraction simultaneously

\nbf{Multi-task training and gradient routing}
Training streams with high input similarity but divergent output targets
risks gradient conflict in the action head.
Our stream-type conditioning embedding mitigates this by routing the
action head to the appropriate output distribution per stream, separating
the output-space conflict while retaining shared upstream representations.
The interaction between streams in the shared backbone is a potential
source of positive transfer (shared retail visual features benefit all
streams) as well as negative transfer (streams that interfere in the
shared representation space).
Per-stream ablation results would quantify this empirically and are
a promising direction for future analysis.

\nbf{Data mixture and the robot-native anchor}
SABER constitutes 85.9\% of the training corpus (44.8K of 52.1K samples), making it the dominant
signal.
The 4,800-sample \texttt{singlepanda} and \texttt{unified\_gr1} robot-native anchor
proved necessary to stabilize early training even at this scale,
suggesting that the presence of general manipulation signal matters for robust convergence.

\nbf{Limitations}
All evaluations are in simulation; real-robot deployment remains future
work.
The body pose retarget stream is substantially smaller (1.2K) than the
other streams; the contribution of additional body pose captures is an
question to be addressed in future efforts.
The stream-type conditioning approach is one design choice---alternatives
such as separate action heads per stream or hierarchical architectures
will be explored in the future.

\section{Conclusion}
\label{sec:conclusion}

We introduced \textbf{SABER}, a high-fidelity retail robotics action dataset
built from approximately 100 hours of natural in-store human activity and 
converted into $\sim$44.8K robot-training samples across three complementary
supervision streams: latent action sequences, dexterous hand retargets, 
and whole-body motion retargets.

The core claim of the paper is that domain-specific deployment in robotics 
is fundamentally a data problem. Retail requires a combination of skill 
diversity, repetition, scene realism, and embodiment-aware action supervision 
that is weakly represented in existing general-purpose robot datasets. SABER 
addresses this by turning real human retail behavior into a reusable robotics 
data layer for post-training, retargeting, and downstream policy adaptation.

Our empirical study demonstrates that SABER-MM multimodal post-training substantially improves retail-task performance: 29.3\% mean success across all ten tasks, with particularly strong results on fridge tasks (82-100\% success) and consistent 13.88\% mean improvement on non-fridge tasks (from 6.0\% baseline). This data-centric approach shows that high-fidelity multimodal action supervision provides complementary signals for retail manipulation learning. The value of SABER extends beyond a single backbone: the same capture pipeline supports latent-action learning, dexterous hand supervision, and whole-body humanoid retargeting, positioning the dataset for broader use in domain-specific robotics deployment across multiple VLA architectures and embodiments.

Future work will complete the full multimodal evaluation suite, expand the
body-pose retarget stream, and validate the dataset on physical retail robots and additional VLA backbones.

\subsection{Future Work}
In this work, we trained the GR00T N1.6 model under several fine-tuning and post-training configurations, keeping only the Flow-Matching Diffusion Transformer unfrozen throughout. While this allowed the model to condition action outputs for the retail domain, the frozen vision encoder limits the acquisition of domain-specific perceptual representations. A natural extension of this work is to train the full model end-to-end on SABER, enabling the vision encoder to adapt alongside the action head and better evaluate generalization in retail settings.
A complementary direction involves replacing the COSMOS-Reason 2B backbone currently used in GR00T with a domain-specialized alternative. Our prior work introduced PRISM \cite{prism-dvu}, a multi-view, multi-capability retail video dataset covering spatial, physical, and action reasoning jointly — capabilities not addressed together in any prior corpus. Fine-tuning COSMOS-Reason 2B on PRISM yielded substantial gains across embodied reasoning, spatial perception, and intuitive physics benchmarks, producing a retail-specialized VLM. Since SABER was designed with modularity and VLM-agnosticism in mind, future studies can use this VLM as a systematic benchmark to compare backbone choices across a range of VLA architectures.

\section{Contributions and Acknowledgments}
\label{sec:contribution}

\subsection{Core Contributors: }
Narsimha Menga, Parikshit Sakurikar, Amirreza Rouhi, Satya Sai Reddy, Anirudh Govil, Sri Harsha Chittajallu, Rajat Aggarwal, Anoop Namboodiri, and Sashi Reddi.

\subsection{Acknowledgments: }
The authors express their sincere gratitude to all members of the DreamVu team who contributed to data collection, annotation, and infrastructure support for the SABER dataset
- Deepu Tiwari, Shaik Arshad, Suresh K, Shrinivas Gone, Sirigiri Manikanta, Farooq Basha Chowdary, Bagara Sai Varun, Yedida Phani Sri Sathvik, Boddu Shyamala, Shanmuki Priya Bala, Nasika Lokesh Vara Prasad, Sukanya Marri, Mahajan Manikanta, Pasunoori Venkatsai, Manisha, K Uday Kiran, Induri Srikanth Reddy, Ayapakola Dwarkesh, Nihasri Gundu, Cholleti Vinay Kumar, Rudra Saikiran, Akuthota Yashwanth, Achintalwar Divya Jyothi, Shaik Mohammad Irfan, Nettem Venkat Ganesh, Anthati Shiva Sai Prasad, Ketham Saikumar, Nenduguri Sai Nithin, Gorli Revathi, Molleti Durga Prasad, Kamatam Shravan, Sayaboina Raju, Gottapu Pravallika, Paras Panse, Rasala Sairam, Cherukuthota Vyshnavi, Jujare Ranjith Kumar.

\bibliographystyle{dreamvu_preamble/assets/plainnat}
\bibliography{references}

\end{document}